\documentclass[pdflatex,sn-basic]{sn-jnl}

\usepackage{amsmath}
\usepackage{amsfonts}
\usepackage[dvipsnames]{xcolor}
\usepackage{bm}
\usepackage{url}
\usepackage{breakurl}
\usepackage{titlesec}
\usepackage{graphicx}
\usepackage{blindtext}
\usepackage{rotating}
\usepackage{titlesec}
\usepackage{subcaption}
\usepackage{hyperref}
\usepackage{pdflscape}
\usepackage{geometry}

\usepackage{pifont}
\newcommand{\xmark}{\ding{55}}%

\begin{document}
\title[\textit{Intelligent Cinematography}: A comprehensive review of AI research for camera-based video production]{\textit{Intelligent Cinematography}: A review of AI research for cinematographic production}

\author*[1]{\fnm{Adrian} \sur{Azzarelli}}\email{a.azzarelli@bristol.ac.uk}

\author[1]{\fnm{Nantheera} \sur{Anantrasirichai}}\email{n.anatrasirichai@bristol.ac.uk}

\author[1]{\fnm{David R} \sur{Bull}}\email{dave.bull@bristol.ac.uk}

\affil[1]{\orgdiv{Bristol Visual Institute}, \orgname{University of Bristol}, \orgaddress{ \country{UK}}}

\abstract{
This paper offers the first comprehensive review of \textbf{artificial intelligence (AI) research in the context of real camera content acquisition for entertainment purposes} and is \textbf{aimed at both researchers and cinematographers}. 
Addressing the lack of review papers in the field of \textit{intelligent cinematography} (IC) and the breadth of related computer vision research, we present a holistic view of the IC landscape while providing technical insight, important for experts across disciplines. We provide technical background on generative AI, object detection, automated camera calibration and 3-D content acquisition, with references  to assist non-technical readers. The application sections categorize work in terms of four production types: \textbf{General Production}, \textbf{Virtual Production}, \textbf{Live Production} and \textbf{Aerial Production}. 
Within each application section, we (1) sub-classify work according to research topic and (2) describe the trends and challenges relevant to each type of production. In the final chapter, we address the greater scope of IC research and summarize the significant potential of this area to influence the creative industries sector. 
We suggest that work relating to virtual production has the greatest potential to impact other mediums of production, driven by the growing interest in LED volumes/stages for in-camera virtual effects (ICVFX) and automated 3-D capture for virtual modeling of real world scenes and actors.  We also address ethical and legal concerns regarding the use of creative AI that impact on artists, actors, technologists and the general public. 
}


\maketitle

\newpage
\tableofcontents
\newpage

\section{Introduction}\label{sec: introduction}
Intelligent cinematography (IC) leverages artificial intelligence (AI) to assist camera-based tasks throughout the media production workflow. For pre-production, this concerns shot planning, for example using a script (\cite{ riedl2010toward, dhahir2022automatic}) or user input (\cite{christianson1996declarative, wu2016analysing}). For production, this includes automated camera-control (\cite{lim2015monocular, coaguila2016selecting, mademlis2019high, bonatti2020autonomous, yan2022real}), methods for content acquisition (\cite{mildenhall2021nerf, pumarola2021d, zhao2022humannerf, legendre2022jointly}) and assisting directors and camera operators (\cite{drucker1995camdroid,de2009virtual, hu2021football, hossain2018exploiting,cao2017realtime, bridgeman2019multi}). For post-production, this involves in-camera virtual effects (ICVFX) (\cite{sharmavisual, kavakli2022virtual, finance2015visual, okun2020ves}), dealing with camera artefacts such as reflections (\cite{wu2022enhancing}), de-rigging and novel viewpoint synthesis (\cite{kerbl20233d, mildenhall2021nerf, yu2021plenoctrees}).

Intelligent cinematography influences the worlds of cinema, gaming and televised broadcasting and streaming. The symbiotic relationship between research and video production has been historically important to the progression of video entertainment. Yet, as this relationship has evolved,  the types of content and the styles of production have increasingly relied on a diverse pool of research across many domains. For example, we find work that automates manual capture processes (\cite{dhahir2022automatic, legendre2022jointly}) as well as work that understands and enhances audience immersion (\cite{hu2021football, wang2013beyond}). This diversity and fluidity of activity makes it challenging to establish a unified definition and description of the IC research landscape.  
Hence there is an evident absence of, and need for, both general review papers and more specialized articles. In this paper, we attempt to overcome these challenges by defining and presenting a structured review on the state and future of IC research. Our definition of IC however specifically focuses on real content acquisition. We exclude research aimed at assisting post-production or automating content generation (e.g., using diffusion models) for two main reasons. Firstly, AI research relating to post-production is already well supported by review papers, like \cite{galvane2015continuity,young2013plans,sharma2023exploring, do2021review}. Secondly, post-production AI typically deals with virtual cameras, objects and worlds, which are typically irrelevant for real content acquisition.

This paper categorizes work by \textit{production medium}. Similar to the definition of an artistic medium, this groups research by the relevant tools and practices involved with the various production types, rather than grouping work by objective. For example, live sport and event broadcasting share many cinematographic practices, so are grouped under the same heading, \textit{Live Production}. This allows us to: (1) further sub-categorize according to related AI research fields and provide a holistic view of the IC landscape while maintaining relevance to each production medium; and (2) include research topics which, while not targeted at creative video production, have direct relevance to it.
The topics covered for each medium are outlined below and are visualized in Figure \ref{fig: taxonomy}.

\begin{landscape}
    \begin{figure}
    \newgeometry{left=-3.6cm, right=0cm, top=0cm, bottom=4cm} 
        \centering
        \includegraphics[ width=1.2\paperwidth]{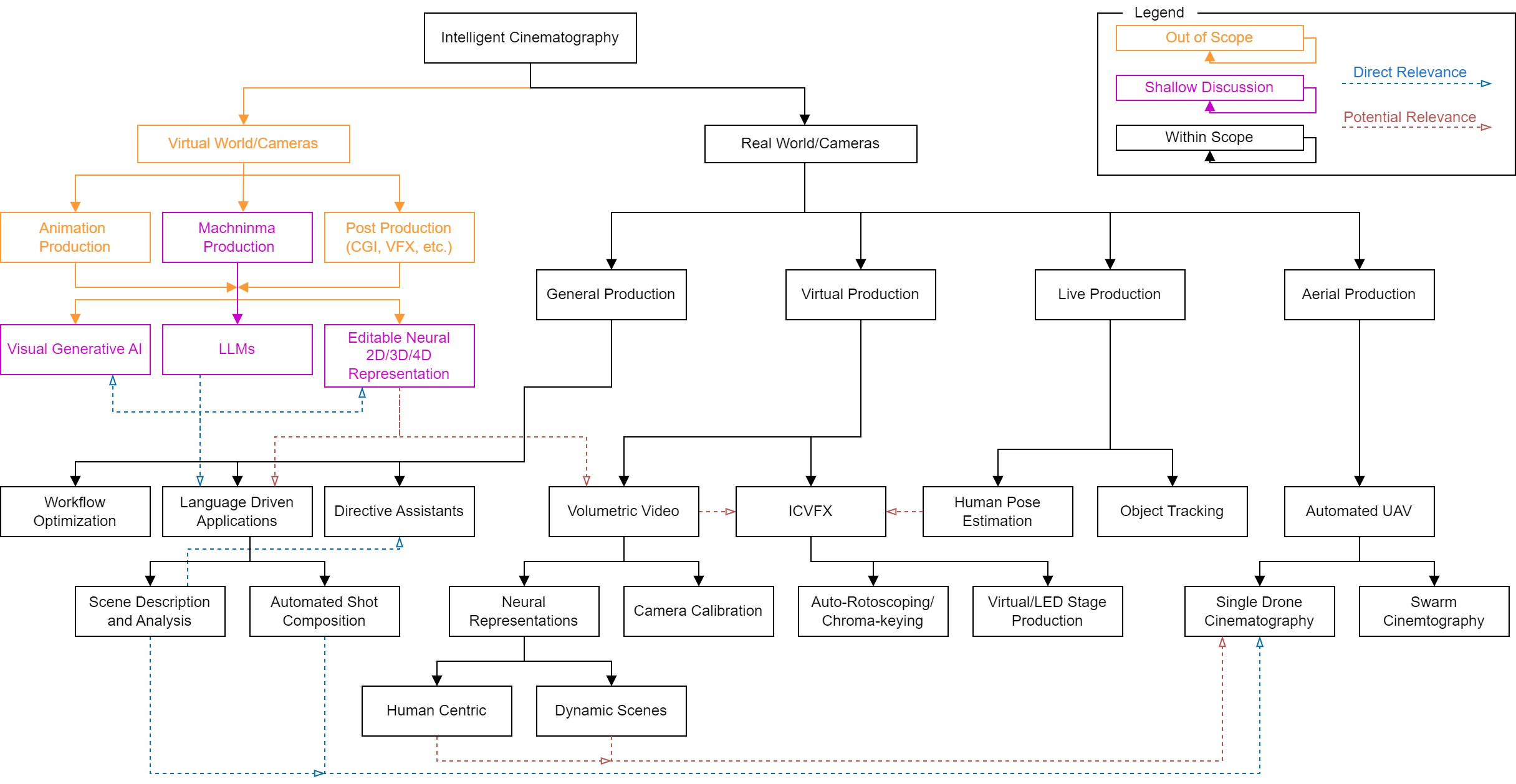}
        \caption{A taxonomy of IC research, including work within and outside the scope of this paper}
        \label{fig: taxonomy}
    \end{figure}
    \restoregeometry 
\end{landscape}

\begin{enumerate}
    \item \textbf{General Production}: Work on human-computer interaction for camera control, automated directive assistance and workflow optimization.
    \item \textbf{Virtual Production}: Work on ICVFX and virtual stage production, and automated 3-D capture of dynamic scenes and human actors.
    \item \textbf{Live Production}: Work on real-time video correction, object tracking and human pose estimation for live broadcasting.
    \item \textbf{Aerial Production}: Work on single UAV and mutli-UAV/\textit{swarm} cinematography, covering topics such as real-time camera planning based on visual aesthetics and safety-first solutions in cluttered filming environments.
\end{enumerate}

\subsection*{Scope and related work} 
Our focus is on camera-oriented research;  hence we do not cover popular creative AI topics such as large language models (LLMs) for script generation or generative AI for video generation. We acknowledge, however, that these may directly impact cinematography in the future. For instance, LLMs may prove useful for text-driven camera control or workflow optimization. Moreover, the rapid progression of generative AI models could deliver high quality video-based methods that offer an alternative solution to automated 3-D capture. As the relevant literature does not yet consider camera-based objectives tied to these technologies we do not investigate these paradigms in the main body. Instead, we highlight potential links in Figure \ref{fig: taxonomy}, provide background on generative AI  in Appendix \ref{sec:mlbackground-autoencoders} and discuss future use cases of these technologies within the relevant subsections.

There exist a small number of articles that review AI in the context of cinematography. For example, \citet{christie2008camera} reviews camera control for computer graphics. This paradigm involves viewpoint computation and motion planning systems, and considers technical constraints such as direct and assisted control, and cinematic constraints (such as view composition and pose). Similarly, \citet{chen2014autonomous} reviews autonomous methods for camera planning, control and selection for multi-camera systems. Furthermore, \citet{qi2010review} reviews camera calibration techniques. Other works look at use-cases for IC such as \citet{galvane2015continuity} for video editing, \citet{young2013plans} and \citet{sharma2023exploring} for gaming and animation, and \citet{mademlis2018challenges} for unmanned aerial vehicle (UAV) control.  Our survey aims to bridge related works to provide a holistic view of the IC landscape so that computing and cinematographic communities mutually benefit from the discussion. 

It is important to note that this review was conducted between September 2022 and May 2024. Considering that many aspects of IC research continues to evolve at a rapid pace, the research represented in this paper will not always reflect the ``current'' state-of-the-art. 
Our intention is instead, to present an organized view of the IC landscape and envision potential areas for future research. In this way, researchers and industrialists alike may benefit, even in the longer term.

The structure of this paper is as follows. In Section~\ref{sec: technological background} we introduce the technical background on AI research commonly found in IC. The main body is contained in Section~\ref{sec: literature review}, where the order of the subsections is reflected in the prior list. For each subsection, we present an overview of the current paradigms and approaches and conclude with a discussion on the current challenges and future works. Additionally, we append Section \ref{sec: social.responsibility} to the main body to highlight the social, ethical and legal challenges associated with IC.
Lastly, in Section~\ref{sec: conclusion}, we offer concluding remarks and outline research that we believe has the most potential moving forward.

\section{Technical Background}\label{sec: technological background}
\subsection{Convolutional Neural Networks}\label{sec:mlintro.cnn}
Convolutional neural networks (CNNs) are the primary technology behind IC. CNNs are a class of neural network that extract the features of an image-type by filtering through a neighborhood of pixels with learnable filter coefficients, to learn common patterns within a set of training images. In the context of IC, CNNs provide backbone architectures for numerous tasks, such as visual generative AI (Appendix \ref{sec:mlbackground-autoencoders}), object detection (Subsection \ref{sec:mlintro.yolo}) and aerial cinematography (Section \ref{sec: aerial production}).

\begin{figure}[ht]
    \centering
    \includegraphics[scale=0.5]{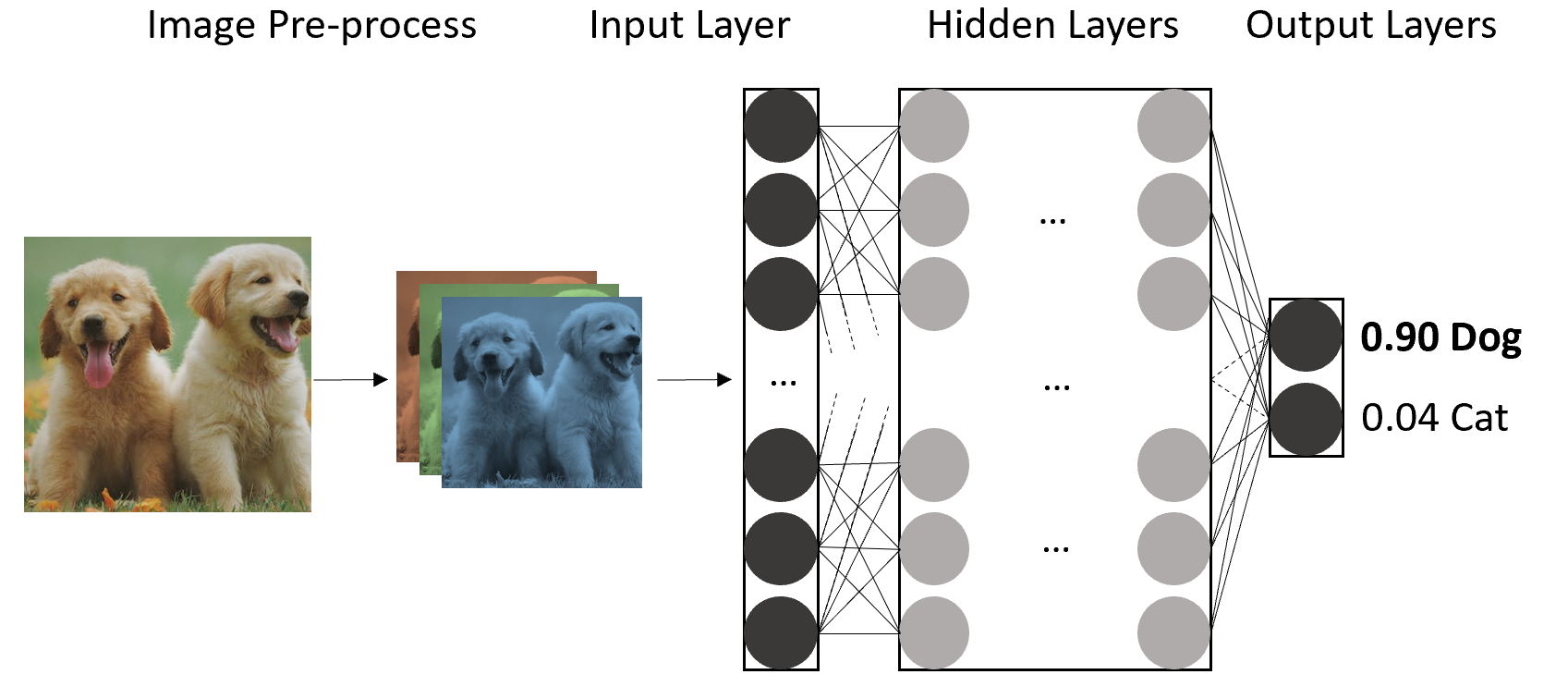}
    \caption{Example diagram of a CNN for image classification}
    \label{fig:cnndiagram}
\end{figure}

A common application of CNNs is in image classification; the basic structure is shown in Figure \ref{fig:cnndiagram}. First, an image is input to the first layer through $N$ channels, where each channel represents a different representation of the image. For example, an RGB image will have three values (red, green and blue) for each pixel, thus $N=3$. The outputs of this layer then feed into a series of hidden layers. The outputs of each layer are called \textit{feature maps}. The last hidden layer transforms these feature maps into feature vectors, so that the output layer can interpret a class probability using the feature vector. An image is then classified if its probability falls within a predefined confidence-interval. In the case of Figure~\ref{fig:cnndiagram}, if our confidence interval is $0.9$, we classify the image as containing dogs.

The hidden layers of a CNN can be flexibly configured. The four components generally employed are \textit{convolution} layer, \textit{activation} function, \textit{pooling} layer and \textit{fully-connected} layer:
\begin{enumerate}
    \item A convolution layer applies an $n \times m$ \textit{kernel} (filter) over each pixel neighborhood from an input channel. To deal with pixels on the perimeter of an image, we generally apply \textit{padding} (the type of padding can vary). Dilated \cite{yu2015multi} or deformable convolutions \cite{dai2017deformable} are also used to modify the structure of the kernel for special cases. In addition, a \textit{stride} can be applied to skip some pixels, i.e. the filter applies to the pixels located $s$ distances from each-other ($s=1$ means all pixels are filtered and $s=2$ means every second pixel in a row and column is taken).

    \item An activation function determines how the features of a layer are transformed into an output. To be used in a hidden layer, the activation function needs to be differentiable and nonlinear, otherwise we would not be able to calculate the gradients of parameters during a backward pass through the network. Rectified linear unit (ReLU) and LeakyReLUs layers are frequently used as they are a differentiable non-linear function, with the added benefit that they are not susceptible to vanishing gradients (where gradients of parameters become too small to make regressive change to the model's parameters). Several types of activation function exist, as discussed in \citet{pragati2022}.

    \item A pooling layer effectively down-samples features, retaining the more important features from a set of feature maps.

    \item A fully-connected layer is one where each node is dependent on all the outputs from the previous layer. This allows the network to introduce a wide range of dependencies between parameters.
\end{enumerate}

\subsection{Object Detection}\label{sec:mlintro.yolo}
In many instances in IC, there is a need not just for automatic object recognition, but also for locating these objects within an image or a video sequence.
For example, in live football broadcasting, there is a need to classify player poses for the purpose of automating camera control. This can assist upcoming phases of a game or event, leading to better organization of shots. Considering the variation of background clutter (e.g., from a crowd, grass, dynamic advertisement banners, etc.), object detection presents significant challenges. Location can be identified by either pixel-level contouring around the object's edge or by a bounding box. The former involves pixel-level classification and segmentation, together called semantic segmentation. In contrast, while the latter incorporates a regression branch to estimate the four corners of the bounding box alongside the classification branch. Object detection using bounding boxes is faster than semantic segmentation and has been extensively utilized in real-time applications, such as sports broadcasting. For example You Only Live Once (YOLO) models for object detection and image segmentation \cite{redmon2016you} only require one pass of a network to infer location. Furthermore, image masking can help YOLO reduce stationary noise \cite{yun2022efficient} by excluding known pixels regions that do not contain relevant information during inference. A real use-case for this was demonstrated by researchers at Kaunas University of Technology, who applied real-time online object detection and segmentation as a means of removing in-camera distractors for broadcasting a soccer game, \citet{ktu2024}.

\begin{figure}
    \centering
    \includegraphics[width=\linewidth]{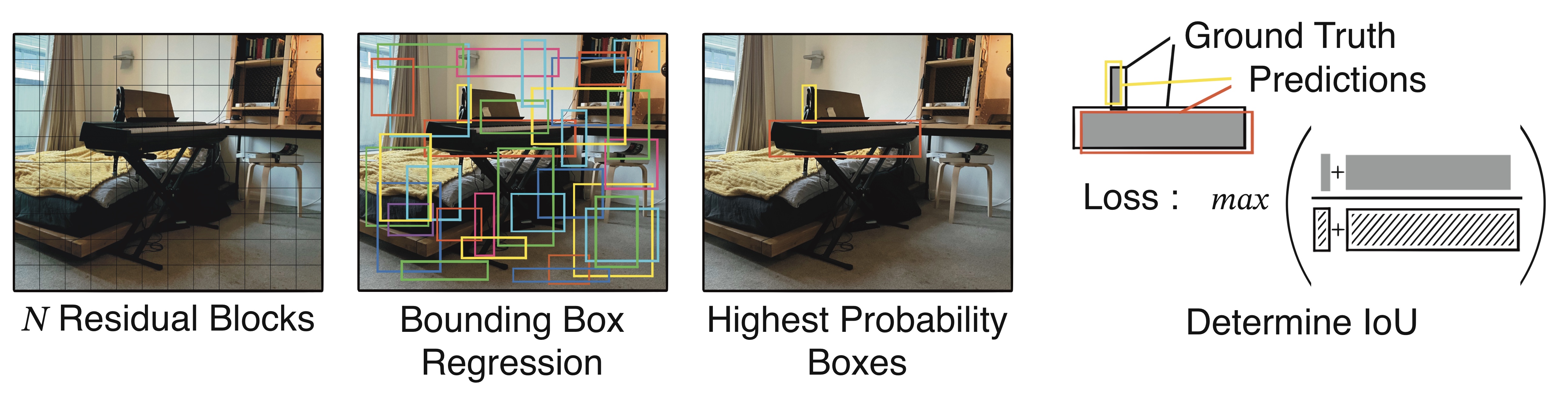}
    \caption{\textbf{Visualizing the YOLO process} with an example of how the IoU is determined. The ground truth is represented by grey boxes and the predictions are the multi-colored bounding boxes}
    \label{fig:yolodiagram}
\end{figure}

YOLO works by dividing a target image/frame into $N$ cells having size $s \times s$ resolution; these are called \textit{residual blocks}. Then, using \textit{bounding box regression} the model estimates which residual blocks are associated with which bounding boxes. The probability of an object being present within the residual block is also predicted. Using \textit{non maximal suppression} the model is able to reduce the influence of low probability scores and determine the bounding boxes with highest confidence; as illustrated in Figure~\ref{fig:yolodiagram}. Following this, to measure the accuracy we take the \textit{intersection over union} (IoU) of a ground truth bounding box and the estimated box to determine how close to the correct bounding box the prediction is. This provides a loss function to use during model optimization \citet{yolo2022}.

Several versions of YOLO have been used in IC applications, the current variant being YOLOv8 \cite{jocher2023yolov8}, which was used in \citet{ktu2024} for detecting and segmenting distractors for soccer games. YOLOv2 supports the prediction of a fixed number objects in a single block, resulting in improved detection of smaller and grouped objects. YOLOv3 builds on this by introducing a model (with higher complexity) that can detect smaller objects and preserve fine detail. YOLOv4 \cite{bochkovskiy2020yolov4} (proposed by different authors) adds to YOLOv3 with \textit{cross mini batch normalization} for higher accuracy and \textit{weighted residual connections} for better convergence during learning. Regarding IC, we find YOLO applied in live broadcasting (Section \ref{sec: liveproduction}) and aerial cinematography (Section \ref{sec: aerial production}).

\subsection{Camera Pose Estimation}
Camera pose estimation approximates the orientation, path and motion (the \textit{extrinsics}) of a camera in 3-D space which, for example, is  useful as a basis for mitigating motion blur caused by the motion of a camera. Accurate camera tracking also enables automated object detection, human pose detection and 3-D capture (photogrammetry). There exist numerous (commercial) solutions for this involving additional hardware (e.g., LIDAR or GPS) for local and global positioning. Though, there is also research on video-based solutions for cases where physical positioning systems are unavailable.

There are three main approaches to pose estimation: (1) Visual Odometry (VO), (2) visual Simultaneous Localization and Mapping (vSLAM) and (3) Structure from Motion (SfM), all of which share common components \cite{taketomi2017visual}. The objective of VO  is to recover a camera path incrementally, optimizing the current pose given the prior set of poses (window bundle adjustment) \cite{guven2021visual}. 
We consider VO as a short-sighted solution as it has trouble linking a poses that intersect previously visited locations, i.e. the new set of frames would be treated as a separate locations. Thus, vSLAM learns an additional global localization map as well as inheriting VO's objective of optimizing the consistency of the local trajectory. The constraint with using vSLAM is that there are limited number of cameras and window bundle adjustments that can be processed. SfM approaches \cite{agarwal2011building, civera20101} overcome this by utilizing measurements taken for every viewpoint/frame to boost reliability and avoid degenerative cases \cite{pollefeys1998flexible}, including those in \citet{molton2000practical, mallet2000position} and \citet{hirschmuller2002fast}. This is a popular means of pre-processing 2D images for 3-D modeling.

Further, discussions on this topic include \citet{saputra2018visual} who compare vSLAM and SfM methods for highly dynamic scenarios, and  \citet{taketomi2017visual} who assess vSLAM and SfM methods between 2010-2016, and \citet{yousif2015overview} who compares state-of-the-art general VO and vSLAM methods for robotic controllers (applicable to automated camera control in IC). There are numerous use cases in IC: for example, in Section \ref{sec: aerial production} we discuss several instances where this research is involved with aerial capture. Additionally, research on camera calibration heavily supports work on automated 3-D/volumetric capture, as the ability to position images in 3-D space is heavily tied to the quality of inverse rendering pipelines \cite{lin2021barf, chng2022garf, barron2022mip}. This discussed further in the following subsection and is revisited in the main body.

\subsection{Automated 3-D Capture}\label{sec:background 3-D modelling}
\subsubsection*{Context}
\begin{figure}[ht]
    \centering
    \begin{subfigure}{\linewidth}
        \includegraphics[width=\linewidth]{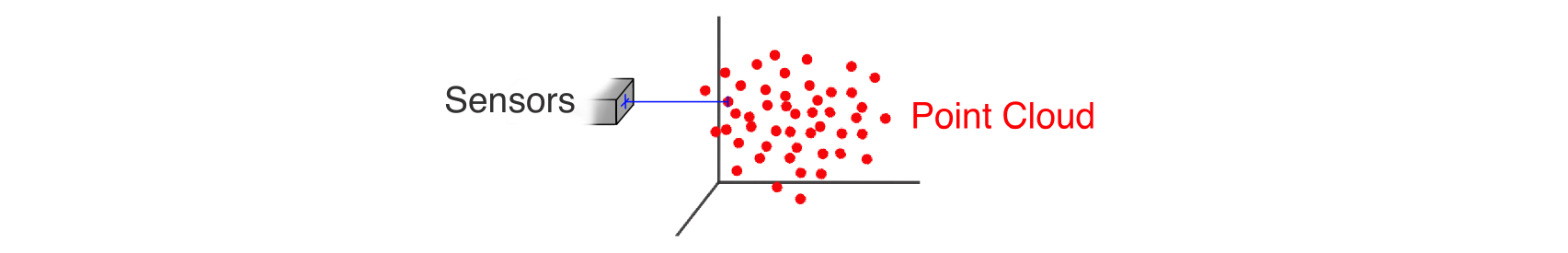}
        \caption{\textbf{Photogrammetry} relies on sensor data (may include cameras) to estimate a point cloud representation of a scene}
    \end{subfigure}
    \begin{subfigure}{\linewidth}
        \includegraphics[width=\linewidth]{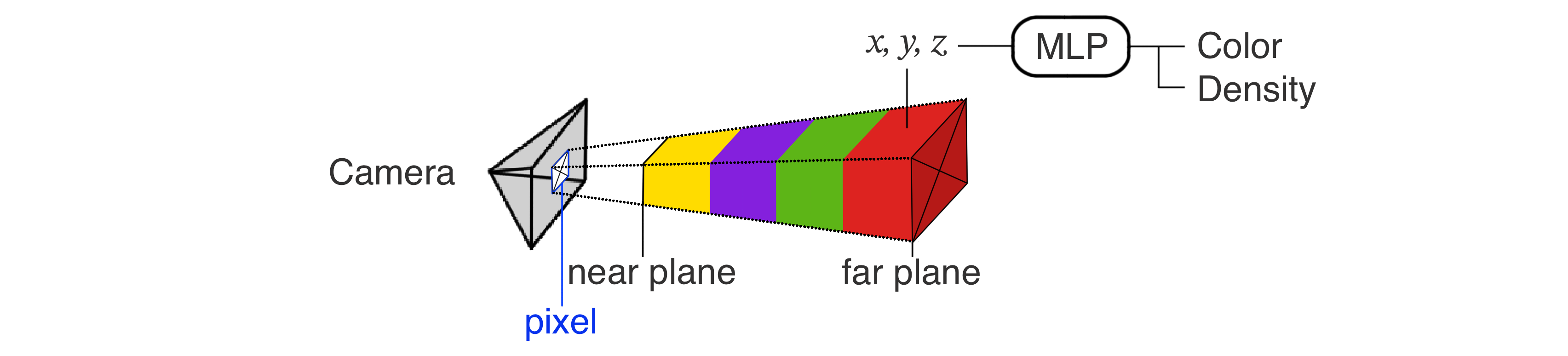}
        \caption{\textbf{Neural Radiance Fields} sample points in space to estimate a color and density value for each sample projected along a pixel-ray}
    \end{subfigure}
    \begin{subfigure}{\linewidth}
        \includegraphics[width=\linewidth]{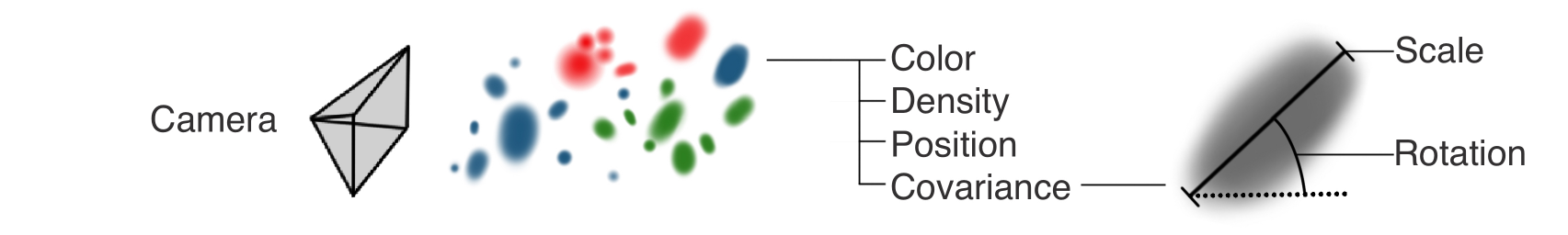}
        \caption{\textbf{Gaussian Splatting} uses a point cloud to estimate the position of samples in space and uses this to estimate the color, density scale and rotation of each Gaussian point}
    \end{subfigure}
    \caption{Simplified representations of current approaches to automated 3-D capture}
    \label{fig: automatedcapture.simple}
\end{figure}

Automated 3-D capture is a popular topic in IC research, whether for analysis or for solving inverse graphics problem. Here, the objective is to discover a 3-D representation using 2-D imagery and/or other sensor data. In Figure \ref{fig: automatedcapture.simple}, we illustrate simplified representations of the current approaches to the problem.

A popular capture method is Laser Informed Photogrammetry (e.g., LiDAR) which creates 3-D point-clouds from measurements of point depth relative to a camera's position. This is now possible  on low cost platforms, such as a mobile phone, thanks to the introduction of mobile LiDAR camera sensors. These methods are promising for capturing enclosed spaces, however challenges remain regarding processing information over greater distances as well as limitations associated with noisy data.

\subsubsection*{Neural Radiance Fields}
Recently, neural radiance field (NeRF) methods have been introduced to solve this problem, whereby the aim is to model the rendering processes involved with automated capture. Unlike Photogammtery, which represent scenes as 3-D structures (mesh, voxel-grid, point cloud), NeRF methods represent scene properties as neural representations which are sampled to interpolate views defined by the camera extrinsics in a virtual space. Given a set of images describing a real environment, NeRF networks can reliably learn the visual features of a scene.

NeRFs represent an important breakthrough for cinematographers, and can be adapted to a range of cinematographic tasks. For example, rendering 3-D from 2-D images can avoid costly re-shoots caused by poor lighting, scenery, weather or acting deficiencies \cite{mildenhall2022nerf, verbin2022ref, yuan2022nerf}. Shots can also be re-worked in post-production, meaning that shot-type, camera jitter, shot duration and focus can also be modified. While the field of NeRF research is relatively new, it has developed rapidly and can now deliver high quality, compact solutions, with the promise of delivering production-ready NeRFs in the near future.

The original NeRF paper \cite{mildenhall2021nerf} demonstrates that, with a good approximation of the camera extrinsic (image position and orientation) and a reliable ray sampler (for sampling points/volumes along a ray) as prerequisites, we can model 3-D scenes that include reflective and transparent materials using a multi-layered perceptron (MLP) neural network. This works by sampling a point along a ray (pertaining to a given pixel in a training image), and using the global 3-D position of that point, $\{x,y,z\}$ and the relative viewing direction of the point, $\{\theta, \phi\}$ as inputs into an MLP that returns an RGB color and an opacity value associated with that point in 3-D space. By accumulating many points along a ray, we can apply a rendering method akin to alpha blending to render each pixel in an image. Through the use of a loss function comparing the rendered image with a ground truth image we can optimize the quality the 3-D scene so that new, unseen views may be generated. Hence, NeRF research falls under the label of \textit{novel view synthesis}. For readers that are interested in a deeper technical discussion, Appendix \ref{app: more-nerf-background} provides a technical introduction to NeRF, how it works and the relevant technical challenges.

\begin{figure}
    \centering
    \includegraphics[width=\textwidth]{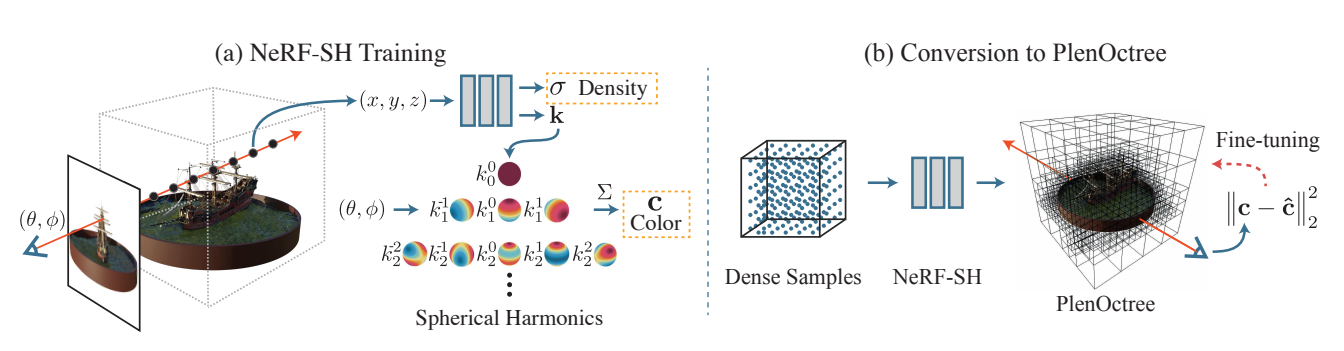}
    \caption{\textbf{Visual comparison of the NeRF and PlenOctrees pipelines} used for neural rendering presented by \citet{yu2021plenoctrees}. Unlike the original NeRF, NeRF-SH uses spherical harmonics to decode visual features into RGB color values. The PlenOctree model refines a 3-D octree representation to model the geometry of a scene}
    \label{fig:plenoctrees}
\end{figure}

While this method works, MLPs take long to train and the quality is often poor and reliant on the number of training images/views we have of the 3-D scene. Subsequently, several approaches have been proposed to speed up NeRF computation: multi-resolution hash encodings, voxel grids and voxel-trees \cite{mueller2022instant, wang2022fourier, yu2021plenoctrees, laine2010efficient}. This is particularly relevant for researchers interested in the dynamic NeRF paradigm, as model complexity increases significantly when modeling 4-D space and time. The PlenOctrees representation \cite{yu2021plenoctrees}, illustrated in Figure~\ref{fig:plenoctrees}, builds off NeRF's ability to functionally represent a view-dependent scene by, (1) representing the structure of a scene as a sparse-voxel octree \cite{wilhelms1992octrees, samet1988overview, laine2010efficient}, and (2) sampling color by involving spherical harmonics. (1) facilitates the rendering step through fast-to-access representations (octrees). Whereas (2) allows us to model view-dependency as octree leaves by mapping the view dependent color values to a sphere's surface and predicting the coefficients of the spherical harmonic equation (with fixed degree) to return a color value where the pixel-ray intersects the sphere. Subsequently, the rendering speed and learning is optimized significantly, and one could argue novel-view quality has improved too. One caveat is the amount of data a representation will consume (in the magnitude of gigabytes), which perhaps is acceptable for those with required hardware, however there exist more affordable explicit options \cite{chen2022tensorf, rho2023masked}. Additionally, there is some difficulty with visualizing unbounded scenes as voxel-grids occupy finite space. \citet{yu2021plenoctrees} overcomes this by using NeRF++ \cite{zhang2020nerfpp} for rendering out-of-bound scenery. This is accomplished by modeling the foreground and background as separate components.

\subsubsection*{Gaussian Splatting}
With similar motivations, 3-D Gaussian Splatting (GS) has been proposed by \citet{kerbl20233d} as a means of significantly reducing rendering time by using: (1) a point cloud representation of Gaussian ``blobs'' with position, covariance, color and opacity properties, which avoids unnecessary computation of empty space present in NeRF models, and (2) \textit{tile splatting} for faster rendering (NeRF typically renders at $< 1$ FPS). 

Tile splatting speeds up rendering through the following steps. We highlight the importance of the sorting approach, (3), as this limits NeRFs from achieving similar computational goals.
\begin{enumerate}
    \item Divide a rendering view into $16 \times 16$ tile
    \item Cull blobs with $< 1 \%$ confidence of intersecting each tile frustum or blobs that fall outside the near and far bounds of the camera
    \item Sort blob depth w.r.t to each tile using a fast GPU Radix sort \cite{merrill2010revisiting}; not per pixel as done in NeRF.
    \item Render pixels w.r.t the sorted blobs for each tile by $\alpha$-blending until the accumulated $\alpha$ for each ray becomes 1
\end{enumerate}
To train a GS model a scene is initialized with a sparse set of point-clouds that are ``densified'' during training by cloning, re-scaling and re-positioning blobs to fit the geometry corresponding to a set of training view/image. Blobs that are essentially transparent (low $\alpha$ contribution) are pruned.

Conclusively, GS maybe selected over NeRFs due to their enhanced computational abilities, and in many cases improved performance. However, existing dynamic GS models are much less compact than NeRF alternatives and are comparatively worse at depth estimation \cite{foroutan2024does}, due to the differences in depth-based rendering approaches Additionally, as GS is a more recent development than NeRF we are yet to see the same attention on cinematographic based tools, such as scene and camera editing or representational transformations such as GS to mesh. However, the rapid development of this research space means that it is only a matter of time before these limitations are addressed.
\section{Intelligent Cinematography in Production}\label{sec: literature review}
\subsection{General Production Applications}\label{sec: general paradigms}
In this section we review research relating to camera management and control, visual analysis, cinematographic assistance and workflow optimization. 

\subsubsection{Computational Language Structures for Scene Analysis, Automated Labeling Schemes and Camera Management}\label{sec:film language}
Cinematographic language is used to communicate the state of production and relevant processes. It enables production staff to effectively communicate and react coherently to unforeseen situations. 
Cinematic theory is often formalized for tasks such as automated camera control and scene analysis. For example, in \citet{jhala2005discourse}  state machines are used to model cinematographic shots for camera planning in a virtual environment. In this context, we discuss current approaches automated camera control that introduce or rely on formal cinematic language for inference; we refer to this as \textit{idiomatic capture}.

A heuristic approach to camera control  was introduced by \citet{jhala2006representational}, based on the \textit{decompositional
partial order causal link planning} algorithm, \cite{young1994dpocl}. This formalizes idiomatic capture  by linking an approximated scene representation to a predetermined set of camera control responses. A conflict resolution algorithm ranks known actions with conditional operators to determine the optimal action and respective duration. To express idioms, \citet{jhala2006representational} distinguishes a set of four requirements to mediate contextual differences between different productions: (1) Story Representation, (2) Real World Representation, (3) Rhetorical Coherence and (4) Temporal Consistency. The first two points describe the physical and contextual landscapes of a set, hence it is necessary to have both geometric (or visual) and semantic representations of a production. The third point expresses the need for rhetorical structure to ensure actions are executed decisively, e.g. a hierarchical-model for selecting shot-types. The final point specifies consistency with regards to the temporal aspect of filming.

\begin{figure}
    \centering
    \includegraphics[width=\linewidth]{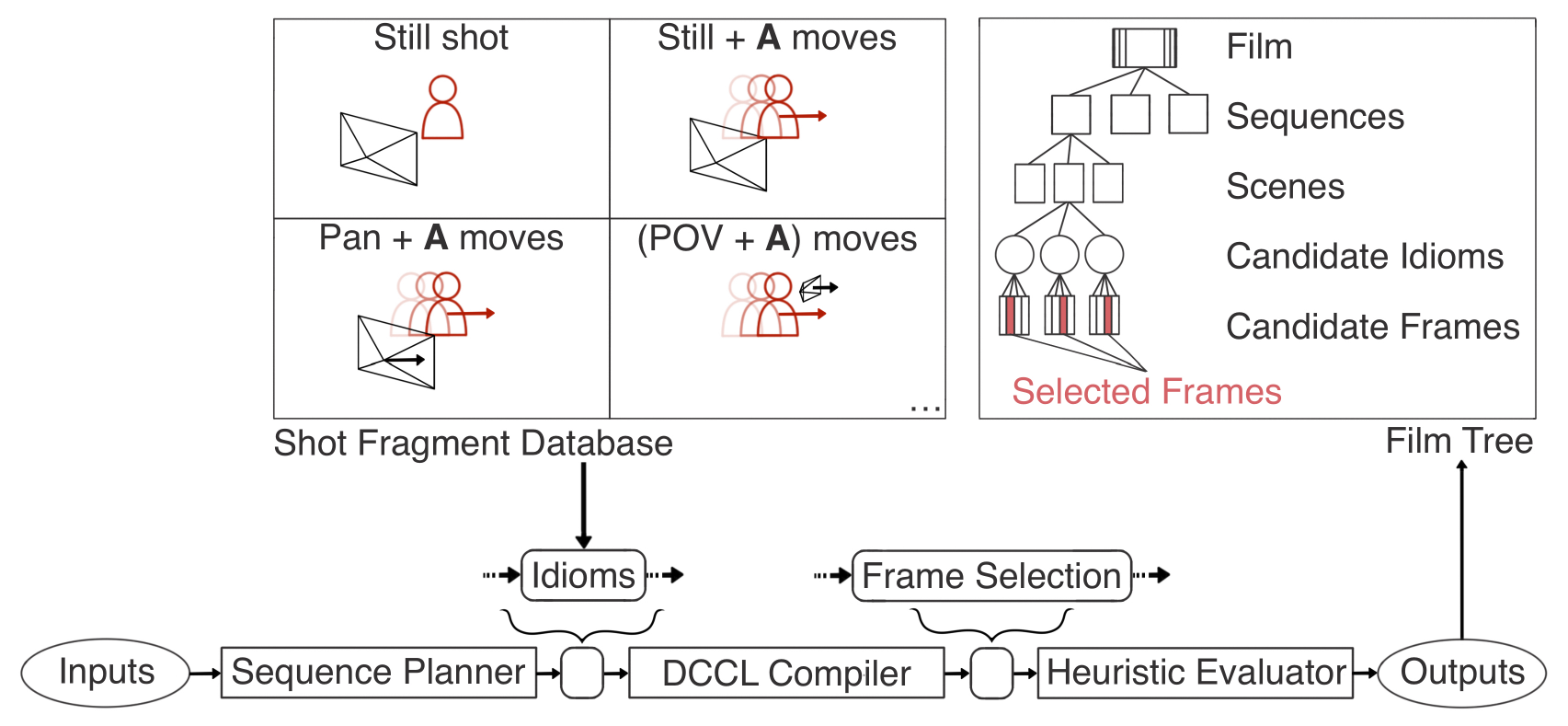}
    \caption{\textbf{Visualizing DCCL for automated camera control:} The three stage pipeline uses idioms defined by camera and actor states as inputs to a DCCL compiler. Selected frames are evaluated w.r.t to the ``film tree''}
    \label{fig: dccl.pipeline}
\end{figure}

Similarly, the Declarative Camera Control Language (DCCL) uses a heuristic-based decision tree for idiomatic capture \cite{christianson1996declarative}, Figure \ref{fig: dccl.pipeline}. This method sets forth a hierarchical structure for automated shot-composition and camera control by breaking down scenes into idiom-specific frame sequences. Here, the relationship of consecutive frame sequences is dependent on temporal links between idioms. A heuristic evaluator then selects a candidate action by scoring possible responses and evaluating decisions via a decision tree.

Idiomatic language structures have also been used for shot labeling \cite{ronfard2022prose} and evaluation \cite{yu2022bridging}. More recently, the Prose Storyboard Language (PSL) \cite{ronfard2022prose} has been proposed as a high-level language for describing shots relative to visual composition and idiomatic camera control. This is demonstrated in Alfred Hitchcock's North by Northwest, where a prompt not only describes the shot-type and foreground, mid and background compositions (separately) but the type of transition as well. As PSL uses \textit{AND/OR decision trees} it can be easily configured to add/modify labels. While the authors have not tested this on an automated labeling scheme, there is cause for such investigation as idiom-based languages might overlook newly adopted idioms and foreign capture practices.

Embedded Constrained Patterns (ECP) is a \textit{dedicated query language}, conscious of physical cinematographic limitations \cite{wu2016analysing}. Unlike PSL, ECP is more comprehensive in its description - labeling montages rather than individual shots. Labels are assigned using the following descriptors:

\begin{enumerate}
    \item \textbf{Framing Constraints}: size of actors and objects, angle of shot, region of actor and object locations and camera movement
    \item \textbf{Shot Relation}: size, angle and region relative to a sequence of shots
    \item \textbf{Sub sequences}: a local grouping of related shot sequences
\end{enumerate}

\citet{wu2016analysing} propose ECP alongside a automated search algorithm for optimal shot-sequence description. The search method is partitioned into two stages, (1) build a cache of valid solutions, and (2) apply a \textit{double recursive depth search} to choose optimal description from the set of valid solutions. During stage (1) valid solutions are separated into three sets: (i) frames that satisfy constraints pre-defined by ECP, (ii) frame couples that satisfy the ECP's relational constraint, and (iii) frame sequences. In the second stage, the search process iterates over all frames in each sequence where subsequent frames are validated as part of the local or global sequence. Alongside the technical implementation of ECP, \citet{wu2016analysing} suggested plans for integration with Unity 3-D to apply ECP to their montages; which we see as a positive contribution towards the production pipeline.

Additionally, languages constructed for automated camera control are not easily comparable. The choice of language structure can vary drastically depending on the application. While there is no current standard for general cinematography, we observe interest from Movie Labs, who look to define a set of computational linguistic structures as a standard for cloud-based workflow optimization \cite{movielabs2030}. The shift to cloud-based computing is indicative of another paradigm of production. Notably, as assets and tools are usually stored or executed offline changes to shared resources can be difficult to monitor on one platform. This could be facilitated by file management approaches which mimic, for example, code-sharing practices. However, this may be difficult to scale for productions requiring a large number of assets or productions which involve external international collaborators who may be un-knowledgeable of standard practices. 

\subsubsection{Directive Assistants}\label{sec:film directive}
Converting most cinematographic concepts into controllers presents a challenging task \cite{christie2009camera}. Notions of shot compositions, shot types and shot transitions are bound by real world problems such as the cost of production, physics and the topology of a set. Thus, AI directive assistants (DAs) have been introduced to alleviate these production challenges. Current IC research addresses tasks such as deriving shot lists, shot plans, optimizing camera placement and controlling robotic camera rigs. 

There is a tendency to use semantic representations of a work environment to provoke DAs. For example, through DCCL, \citet{christianson1996declarative} demonstrates a heuristic approach that relies on idiomatic-based practice for decision making on shot composition. Similarly, \citet{he1996virtual} discuss a  different heuristic approach, utilizing a set of \textit{finite state machines}\footnote{A state-based control architecture for decision making. If approached intelligently (e.g., state labels represented by a bit-string), logical optimization can be applied through \textit{state transition tables} and \textit{state maps}/\textit{implication charts}. More on this here: \url{https://inst.eecs.berkeley.edu/~cs150/sp00/classnotes/katz-ch9-mod.pdf}.} 
to handle idiomatic camera actions. Both of these methods use virtual simulations with user-led events to demonstrate the ability of their tools to derive an idiomatic responses in a short amount of time.

There are more novel approaches to the problem of deriving a shot-list, such as \citet{de2009virtual} who employ an architecture that simulates four critical filming roles: (1) a Script Writer who observes the context of a current scene and sends information to (2), (2) a Screen-orgrapher who configures the staging of actors and objects for dramatic effect and passes this information to (3) and (4), (3) a Director who extracts important information and uses multiple support vector machines (SVMs) to make decisions on selected shots, and (4) a Cameraman who follows idioms for shooting. This method compounds three 2-D feature matrices into a high dimensional feature space - selecting the most relevant SVM from a pre-trained set to classify the optimal shot position and viewing angle. SVMs work by using kernel maps to remap all features into a high-dimensional matrix. This is done under the assumption that classes of features are not linearly separable until they are compounded into a high-dimension. In practice, normalized positional values represent the environment features, while the perceived emotional state of each actor is represented by the actors' features, and the actor who is the principal focus is represented within the scene's features. These are used to for selecting the right SVM to estimate the optimal camera location and angle given the context of the scene. As SVMs may be considered old w.r.t to the current literature, there are likely more suitable approaches for modeling the environment, actor and scene features to classify the optimal camera angle and location from a pre-defined set. For example, with a large enough dataset a deep neural networks will generalize better. There are also faster and simpler approaches such as using K-nearest neighbours on the environment, scene and actor features.

\begin{figure*}[ht]
    \centering
    \includegraphics[width=0.7\textwidth]{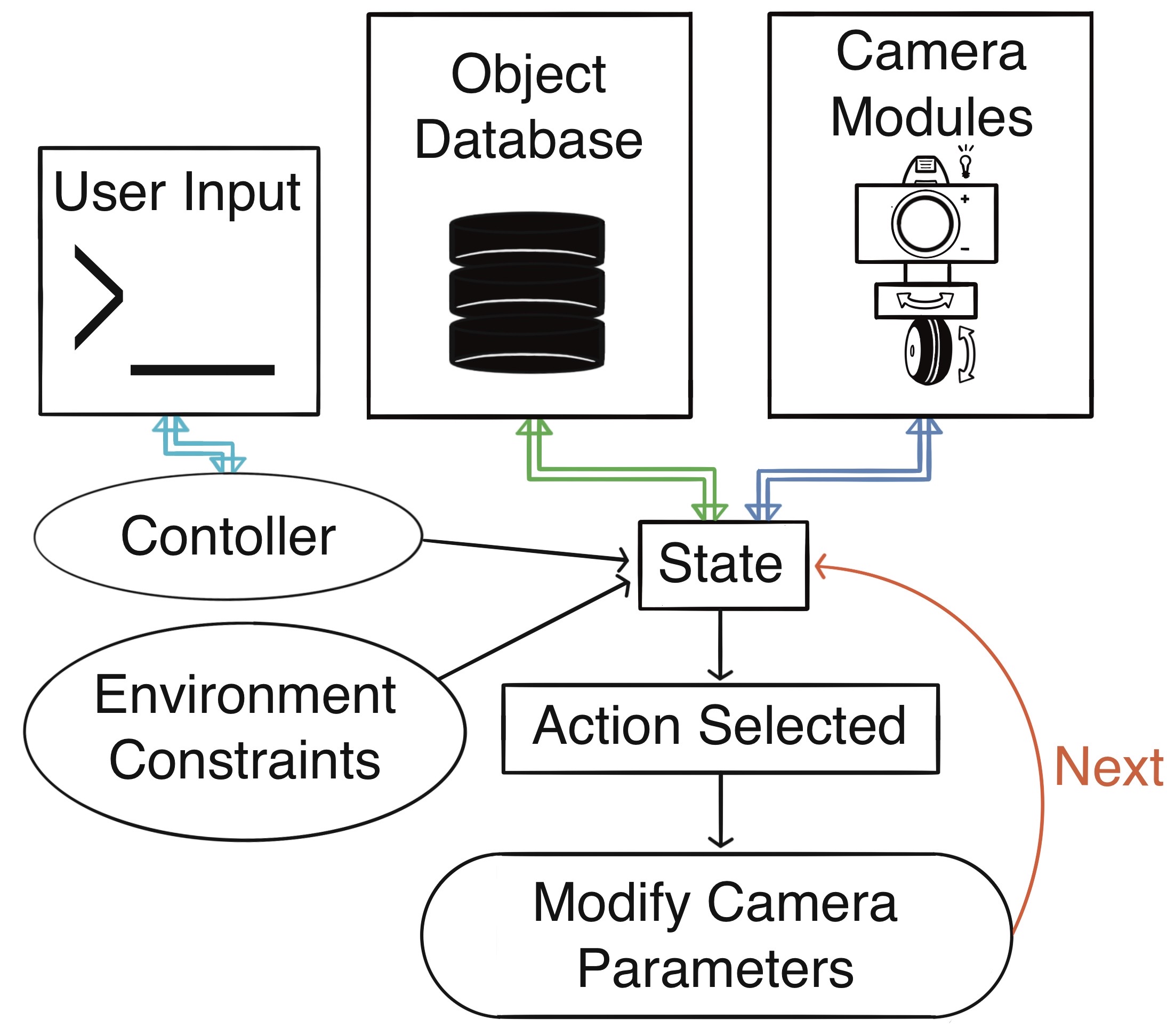}
    \caption{\textbf{Visualizing CamDroid for camera control:} A state-based architecture that relies on user inputs and environment constraints to modify camera parameters for an action sequence}
    \label{fig:camdroid}
\end{figure*}

Alternatively, CamDroid is a well known state-based camera control architecture, that was introduced by \citet{drucker1995camdroid} as illustrated in Figure~\ref{fig:camdroid}. This method uses information on the local real environment to inform a sequence of actions to be taken for optimal capture. Whereby, different visual queues or user prompts will trigger a sequence of camera actions. Conditional functional frameworks like CamDroid are a classical way of handling automated camera control and are powerful for capturing real-time time-dependent action.

Despite the sparsity of DA research, there are recent works such as \citet{stoll2023automatic} who investigate shot selection with a multi-view camera setting for filming theatrical performances. The authors record performances from multiple views in 4K and by cropping the high-resolution frames into lower resolution frames, a set of camera actions is derived. Subsequently, skeletal and facial poses are estimated for each actor in the set of cropped videos. These are used in an automated editing script where, for example, moving lips may be used to detect an actor as the principle focus of the scene, so the relevant clip (i.e. a selected view and camera action) are selected for a given time frame. Preceding \citet{stoll2023automatic}, prior works have tackled zoom and crop methods in different settings \cite{gandhi2013detecting, gandhi2014multi, rachavarapu2018watch} such as \citet{kumar2017zooming} who displays the set of cropped shots in a split screen for shot selection.

The principal areas for investigating DAs: through idiomatic language, classification and state based approaches, are not novel by today's standards of AI research, but are nonetheless capable of achieving reliable idiomatic camera actions in response to user input. However, because the discussion on concurrent methods is limited by the sparsity of recent work, we have yet to see promising work that moves away from offline automated DA and towards a live setting, such as in sport broadcasting. Despite this, numerous related fields approach similar paradigms. For instance \citet{besanccon2021state} discuss state-of-the-art approaches for interacting with 3-D visualizations. The authors outline numerous areas for improvement, particularly surrounding the tools and error metrics used to understand and evaluate human-computer interaction. Skeletal estimating and facial recognition, such as in \citet{stoll2023automatic}, are not sufficient to gain the context of a scene. Likewise, \citet{ali2008applications} discuss applications relating to camera control in virtual environments, which also relates to the discussions in Sections \ref{sec: camera calibration} and \ref{sec: learning dynamic 3-D representations}. 

In the following subsection we present recent approaches to workflow optimization that encompasses aspects of the DA paradigm, such as language-driven shot selection.

\subsubsection{Workflow Optimization and Automated Shot Composition}\label{sec:optimising workflow}
Pre-producton and production workflows will vary drastically as a result of budget constraints, delivery deadlines and creative objectives. Despite this, there are aspects to film-making which remain constant, including set design, the object and background staging processes, and film capture and editing. In industry, notable efforts from Movie Labs (discussed in Section~\ref{sec:film language}), illustrates the potential of a cloud-based platform for streamlining production, entailing new computational language structures, methods for cloud-security and collaboration workflows. Movie Labs have produced a set of white papers \cite{movielabs2030} detailing their ambitions over coming next decade.

Machinima Production (MP) for IC leverages intelligent script writing to generate camera poses for shots (tested on a virtual scene)\footnote{Further discussions: \citet{riedl2010toward} and \citet{elson2007lightweight}.}. Rather than automating the entire production process, we see potential in adapting MP tools to attend to the general scope of workflows (not necessarily for computer generated animation).
For example, the GLAMOUR system \cite{dhahir2022automatic} utilizes \textit{natural language generation} informed by cinematographic idioms, to produce a movie-like composition of shots, with comprehensive descriptors of the scene from still images. The outcome being several short documentary-style productions. GLAMOUR is a multi-objective heuristic approach for \textit{attention optimization}. \citet{pui2021artificial}, thus as animated actors are directed (by an AI) during a scene, the optimal choice of camera and transition are determined. As discussed by \citet{tan2018psychology}, the level of potential stimuli one can experience from a film far surpasses the complexity of audience attention. Thus, GLAMOUR could be extended to include other processes involved in automated shot composition, such as \textit{kernalized filter control} (KCF), presented by \citet{henriques2014high} and used for actor detection and framing in automated aerial content acquisition (discussed in Section \ref{sec: automated single uav}). Additionally useful, cinematic motion style transfer in \citet{wang2023jaws} uses 3-D representations generated from real images and matched to a shot sequence from a scene in a given movie. This could be paired with attention optimization to improve upon selected idiomatic shot to transfer, from which we could derive more meaningful camera poses and motions.

\begin{figure}
    \centering
\includegraphics[width=1.1\textwidth]{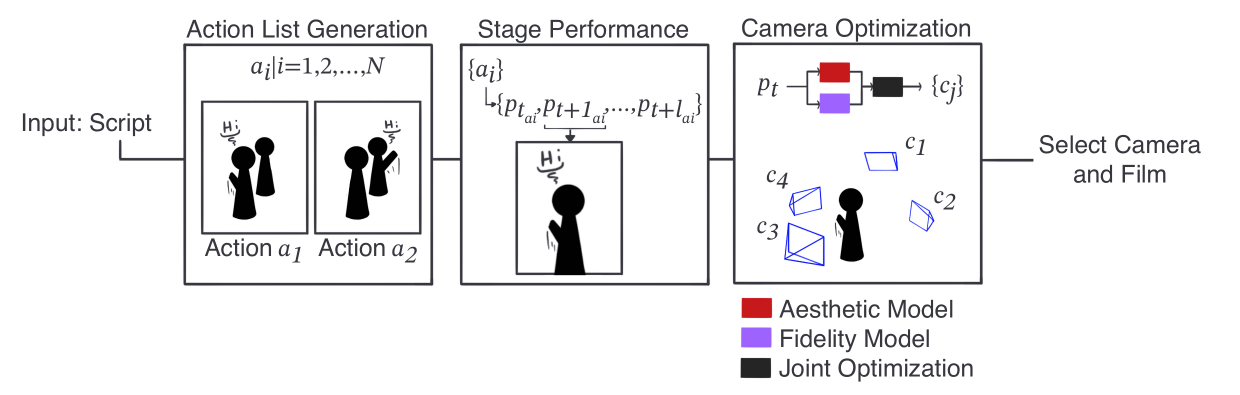}
    \caption{\textbf{\textit{Text2animation} for shot selection} discussed in \citet{yu2022bridging}, where $c_j$ is the camera selected for capturing action $a_i$ during the performance $p_k$ and relies on jointly learning an aesthetic and fidelity model}
    \label{fig:autocinema}
\end{figure}

A semi-automated method introduced by \citet{yu2022bridging} was designed to handle object staging, automated camera positioning and shot composition, from an annotated script. The framework shown in Figure~\ref{fig:autocinema} optimizes camera parameters dependent on a generated sequence of actions for each present character. An action list is transformed to a stage performance, where each action corresponds to a movement (time period) in a scene. This automates the scene scheduling process. During the camera optimization step, ``aesthetic" and ``fidelity" models are jointly applied to a performance, to determine the optimal camera to use at each time step. The aesthetic model analyses six factors for camera planning: character visibility, character action, camera configuration, screen continuity (relative to character position), moving continuity (accounting for on-going changes between movements) and shot duration. The fidelity model first assumes that a mathematical model can approximate the relationship between a script and a generated video. To accomplish this, the model uses the \textit{global vector for word} (GloVe) embedding model \cite{pennington2014glove} to generate text from the generated video, then analyses the similarities between generated script and target script\footnote{The final result is visualized here: \url{https://www.youtube.com/watch?v=0PUdV6OeMac}.}. A GloVe embedding is vector representation for words that are trained using the global co-occurrence word-word of word pairs/groups. Thus, sub-structures of the vector space can define synonyms (proximal parallel vectors) and canonical structures (vector paths). This is similar to the latent variables for text-based generative AI, discussed in Appendix \ref{sec:mlbackground-autoencoders}.

Overall, both of these methods produce naive approaches to workflow optimization - constrained both by cinematographic understanding and current technological capability. For example, the crux of the evaluation mechanism in \citet{yu2022bridging} depends not only on the assumption that a fidelity model can be constructed, but also the ability to re-translate a video into text in a manner appropriate for evaluation. The reliance on this chained processes reduces the effectiveness of this model. Though as \citet{yu2022bridging} concludes, there is still much to accomplish in optimizing production workflows. For example, a further extension of method like GLAMOUR could involve automating the generation of annotated shots or for extending story-board process. This could prove beneficial for productions with tighter budgets and/or deadlines. Additionally, an improved method of evaluating such models is needed, as we have doubts as to the validity of text-based testing for cinematographic models.

Considering that workflows are often fragmented due to reliance on external collaborators, research focused on supporting collaboration seems appropriate. For example, a model could be tasked with acting as an administrator for a shared resource - deciding if committed changes should be accepted given some camera-based constrains. Leaning into ICVFX, cameras such as the Axibo PT4\footnote{Product information: \url{https://www.axibo.com/product/pt4}.} motion controlled slider host a cloud-based Unreal Engine workflow and directly attend to the shared resource paradigm. Use of such tools could be extended to investigations projecting dynamic assets \cite{dang2019deep, lan2022vision, ji2020survey} and validating the in-camera (artistic) composition given a feature-based target \cite{lee2018photographic, wang2023jaws}.

\subsubsection{Challenges and Future Work}\label{sec: general.challenges}
One of the challenges with introducing novel AI into the production process is how to test and validate it. Unfortunately, real production environments are not only challenging to access but also difficult to control. Therefore most proposals favor testing using simulation via virtual environments. Clearly, there is disparity between a semi-informed simulation and a real use case. Going forward, we need to acknowledge the copious benefit of experimental productions \cite{rob2013no}. Experimental productions could become a platform for testing IC tools, whether this is to accomplish menial tasks so artists can focus on creative experimentation or to explicitly support one's creative expression. Examples of this already exist, such as Lux Aeterna who produced a sci-fi movie, \textit{RENO}, as part of the MyWorld Strength in Places Programme, as a means of exploring AI tool-sets in visual effects, \citet{rob2024pioneering}. This RENO production was used to explore the use of AI in post-production, however such work could be elevated to production-oriented tasks. Alternatively, we can envision using 3-D reconstructions of real scenes, discussed in Section \ref{sec: virtual production}, to achieve more reliable tests on real scenarios. This would provide more relatable results for cinematographers that intend using IC to direct cameras in real environments, as well as those that wish capture cinematic footage using 3-D reconstructions.

With respect to supporting the cinematographic language, there are many considerations. For example, the application of language can be beneficial for semantic script analysis/generation \cite{hladun2021intelligent, dharaniya2023design, martinez2019violence} but it may also act as a form of communication between production staff and AI interfaces \cite{christianson1996declarative, yu2022bridging}. We found PSL to be particularly interesting as it is represented as mid-level language. This is beneficial as it reduces the reliance of solving natural language processing (NLP) problems (e.g., a sequence2sequence encoding/decoding problem), which can be troublesome when confronted with complex shot descriptions or tasked with describing subjective cinematographic observations. However, we must acknowledge the progress made in NLP and semantic analysis research over the last decade \cite{cambria2014jumping, chandrasekaran2021evolution}, which could lead to better proposals for learning linguistic structures for cinematographic production. For example, riding off recent breakthroughs in NLP \cite{radford2018improving, mhlanga2023value} researchers could explore LLMs for camera control and workflow optimization tasks, reliant of course on establishing reliable sources of data. For example, for semantic shot analysis one could propose a comprehensive dataset and method of labeling shots using PSL to reduce the complexity of natural language structures for word-embeddings.

Otherwise, DA research provides a platform to exploit a range of AI controllers. These can be adapted for specific purposes and architectures are often implemented as modular pipelines. However, evaluating performance in-the-wild is limited by access to real sets, production staff and the limited number of benchmark datasets and models. This indicates that the state of DA research still has a considerable journey before solutions can be applied on a commercial scale. We hypothesize that as production is further facilitated in other areas, such as improved camera control \cite{amerson2005real, christie2008camera, jhala2006representational}, DA could be facilitated by controlling a set of automated tasks on a higher-plane of abstraction, perhaps through the use of semantic query languages for resource descriptive frameworks (RDF) \cite{creighton2006software, haase2004comparison} or DCCL and PSL.

With respect to workflow optimization strategies, we must examine the social benefit of MP research. As discussed in Section~\ref{sec: introduction}, the aim of IC is to better facilitate artistic film-making. However, automating the entire process (from `text-to-animation') may leave little room for creative input \cite{jason2022can, ploin2022machine, ox2022art}. MP could be used to support story-boarding and other pre-production tasks. Taking this further, one could formulate a storyboard as a set of animated clips to make decisions on shot composition. Thus future work could look at optimizing shot composition, such as for style transfer \cite{wang2023jaws} or attention optimization \cite{valuch2014effect, piao2019depth}.

\subsection{Virtual Production}\label{sec: virtual production}
In this section we focus on virtual production using real cameras. This is important as it delineates from classical virtual production, used in animation or gaming. Two notable differences exist: firstly, compositing virtual scenes for animation is less challenging than compositing virtual assets in a real scene for an IC production as nuances in lighting, coloring and perspective between real and virtual assets require attention. Secondly, real cameras are limited by physical and fiscal constraints, like set topology, additional hardware and technical skill for achieving specific camera motions. These matters are trivial to accomplish with a virtual camera that has 6 Degrees of Freedom (6DoF). 

These differences underpin the challenges of research tied to virtual production for IC. Hence, in this section we review research relating to ICVFX and LED Volumes - touching upon works that investigate re-colorization and \textit{image-based lighting} (IBL). We also discuss research focused on synthesizing virtual replicas of real actors and scenes through NeRFs, prefaced in Section \ref{sec:background 3-D modelling}. This looks at removing the physical and fiscal limitations of using real cameras for content acquisition, allowing users to re-capture real scenes in the context of a virtual environment/engine, as is achieved with classical virtual production.

\subsubsection{ICVFX and LED Volumes}\label{sec: icvfx and vsp}
\begin{figure}[ht]
    \centering
    \includegraphics[width=\linewidth]{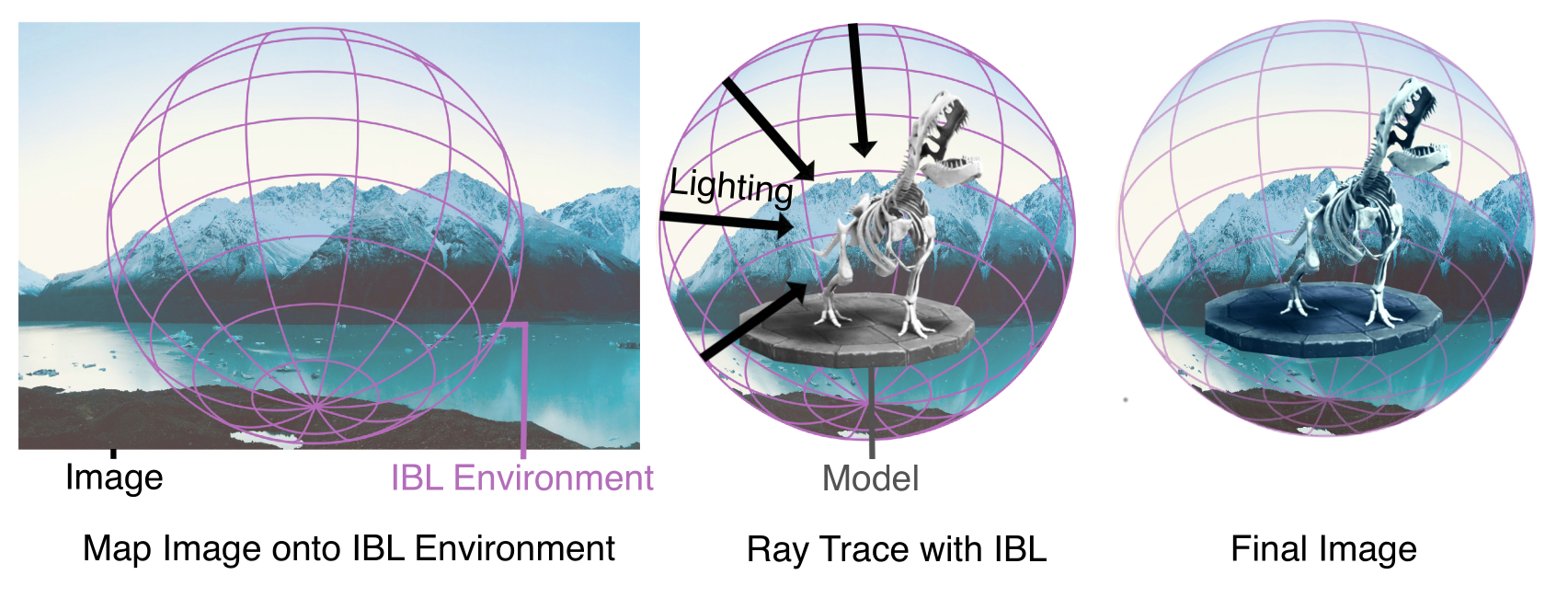}
    \caption{\textbf{Image based lighting process for compositing:} IBL maps an environment image onto a 3-D primitive (e.g., a sphere). The IBL environment is treated as a light source and using ray tracing a model is composited.}
    \label{fig: ibl-figure}
\end{figure}

\begin{figure}[ht]
    \centering
    \begin{subfigure}{\linewidth}
        \includegraphics[width=\linewidth]{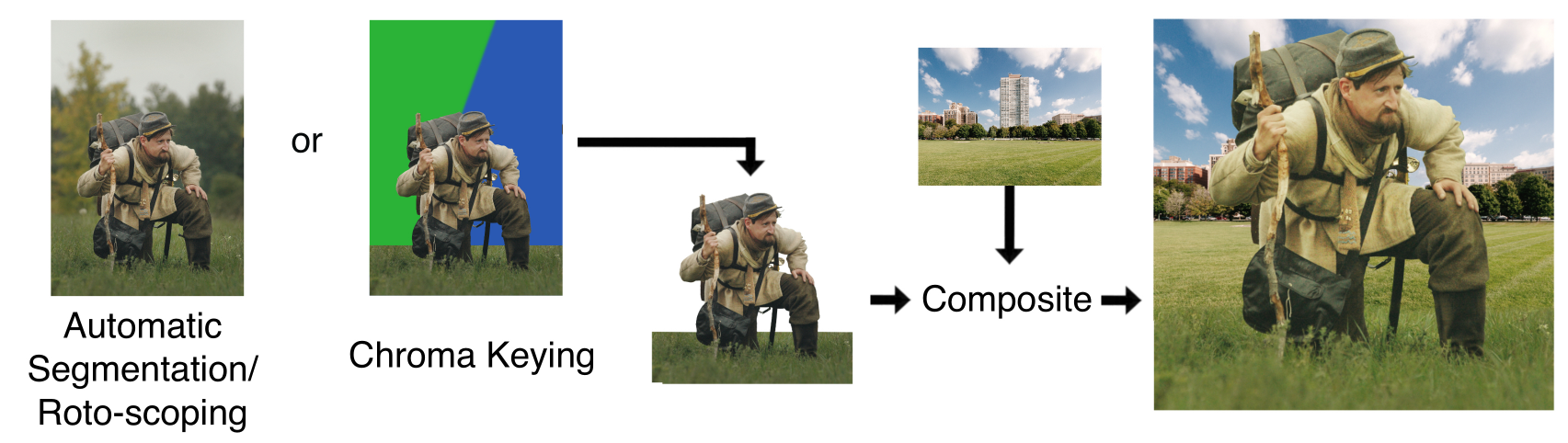}
        \caption{Foreground separation with automated segmentation or chroma screens.}
    \end{subfigure}
    \begin{subfigure}{\linewidth}
        \includegraphics[width=\linewidth]{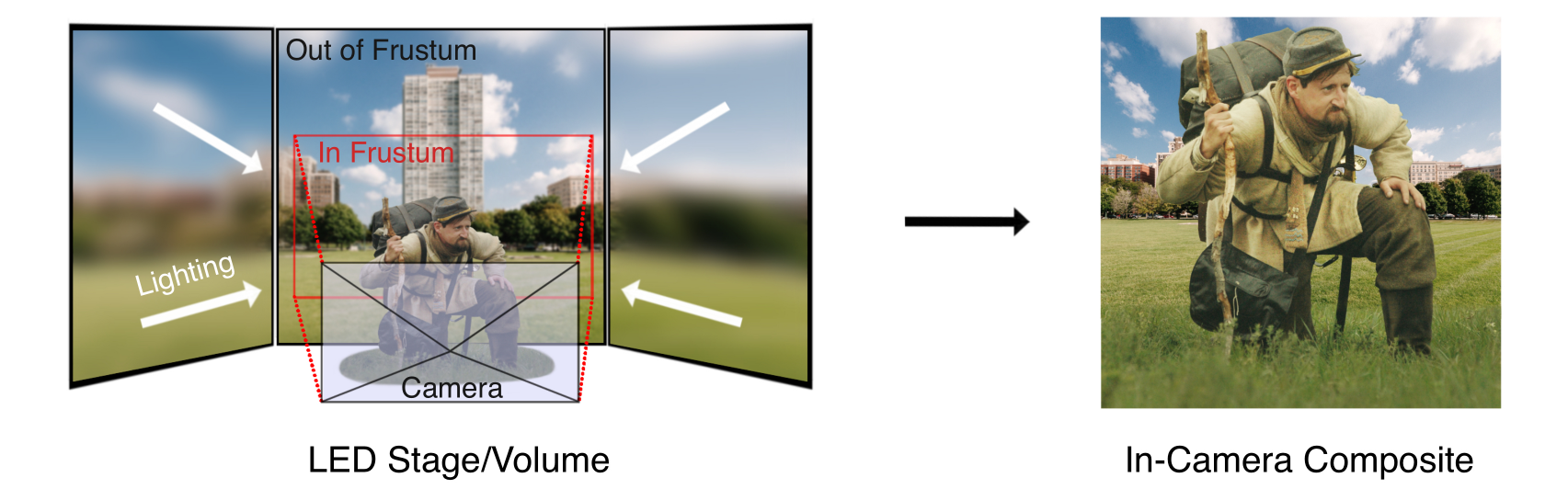}
        \caption{In-camera compositing with an LED stage}
    \end{subfigure}
    \caption{\textbf{Approaches to in-camera compositing:} using automatic segmentation, chroma keying and LED stage/volume}
    \label{fig: icvfx-figure}
\end{figure}

Numerous scholarly works have undertaken the task of dissecting visual effects (VFX) and in-camera applications within the realm of cinematography. However, the relevant use case(s) for IC is vaguely defined. Considering a purely cinematographic perspective, ICVFX usually involves CGI and/or compositing techniques that are executed in real time, providing cinematographers with a live feed of how virtual effects will appear relative to the real scene set-up. This can support production in numerous ways, such as providing an indicator for poor lighting or as a way to pre-visualize compositing and VFX to ease post-production challenges. Here, technical practice mainly involves distinguishing foreground and background elements \cite{sharmavisual} as well as modifying lighting and color \cite{bengtsson2022image}. \citet{sharmavisual} briefly presents \textit{chroma key} and \textit{roto scoping} as paradigms, with a focus on the former; as illustrated in Figure \ref{fig: icvfx-figure}(a). While, \citet{bengtsson2022image} presents how IBL, a well understood method in VFX in Figure \ref{fig: ibl-figure}, has evolved into a lighting workflow surrounding LED video screens for driving in-camera relighting and colorization, Figure \ref{fig: icvfx-figure}(b). This hints at two distinct use cases for IC research:
\begin{enumerate}
    \item Automatic foreground segmentation\footnote{Here, the terms segmentation and roto scoping can be used interchangeably.}: (i) with and (ii) without chroma screens (i.e. blue/green screens)
    \item Automatic IBL and re-colorization
\end{enumerate}

It is generally known that use case (1) is readily accomplished by using chroma screens (1.i) and keying out the color corresponding to the screen. However this introduces issues for color-based image segmentation. \citet{cheng2001color} touches upon relevant color-based image segmentation techniques angled towards the wider AI audience. The authors discuss solutions such as histogram thresholding \cite{littmann1997adaptive}, using binary trees to store data intensive 3-D color spaces \cite{schacter1976scene, sarabi1981segmentation}, region-based methods \cite{tremeau1997region, cheng2000hierarchical}, fuzzy techniques \cite{huntsherger1985iterative, tominaga1986color} and neural networks \cite{huang1999pattern, campadelli1997color}. Consequently \citet{cheng2001color} highlight problems with shading (e.g., shadows and highlights) and texturing. The most apparent case of this is interfering light-bounce from the chroma screen. This leads to subject specific approaches, for example \citet{sigal2004skin} approaches real-time skin segmentation for video opposed by time-varying illumination. \citet{sigal2004skin} optimizes a second order Markov model to predict a skin-color histogram over time. Additionally, chroma keying restricts the ability to present colors similar to that of the chroma screen.

More recently, we have observed approaches attending to case (1.ii). This is a popular paradigm outside of cinematography whereby we see emerging research such as the Segment Anything Model (SAM) \cite{kirillov2023segment} - a one-click solution to general image segmentation. Angled towards cinematography we find work such as Roto++ \cite{roto++2016} - a rotoscoping tool with the ambition of respecting the artists' requirements. Roto++ improves upon traditional interpolation techniques by combining a real-time appearance tracking and a novel shape manifold regularization process (built on the Gaussian process latent variable model (GP-LVM) \cite{lawrence2005probabilistic}). Subsequently, within a sequence of frames the method (1) predicts the change of a shape manifold and (2) identifies which next keyframe needs to be manually roto scoped.

Concerning case (2); while IBL is classically a 3-D rendering technique, it can be applied to real scenes where the lighting set-up can be readily changed \cite{debevec2006image}. \citet{ren2015image} and \citet{wang2009kernel} discuss approaches that reconstruct the light transport matrix (LTM)\footnote{This defines light interactions on object surfaces.}. \citet{wang2009kernel} classifies the approaches into three categories: (i) \textit{brute force} \cite{o2012primal, hawkins2005dual}, directly modeling the LTM, (ii) \textit{sparsity based} \cite{garg2006symmetric, masselus2003relighting, reddy2012frequency}, modeling a set of basis functions under the assumption that each row of the LTM can be linearly approximated, and (iii) \textit{coherence based} \cite{o2010optical, wang2009kernel, fuchs2007adaptive}, analysing the coherence reflectance field to acquire the LTM. The limitation with these methods is that they require multiple images under varying lighting condition, meaning for video this problem is more challenging to address.

\begin{figure}[ht]
    \centering
    \includegraphics[scale=0.8]{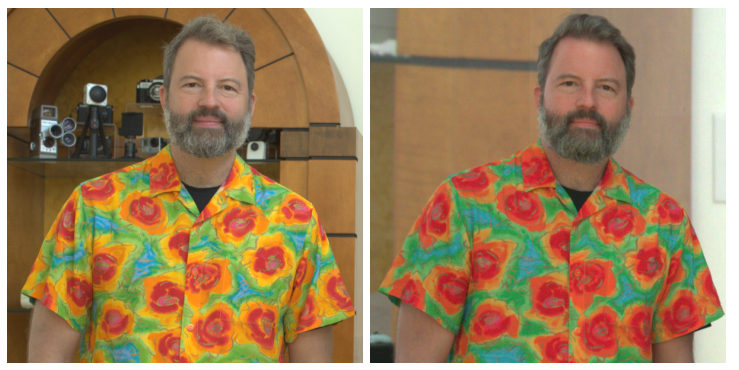}
    \caption{Results from images taken in natural lighting (left) and configuring the surrounding LED panel stage/volume for in-camera lighting (right) provided by \citet{legendre2022jointly}}
    \label{fig:legendre2022example}
\end{figure}

Interestingly, focus on cases (1) and (2) has shifted towards using LED panels to project virtual backgrounds as a practical solution. For case (1) foreground-background separation is made easy by replacing chroma screens with virtual backgrounds displayed live on interconnected LED panels. While more costly and energy intensive, it reduces the workload for post-production and avoids the need to roto scope. Unfortunately this means the need for research in this area is minimal and shifts toward supporting computation (e.g., reducing energy consumption). For case (2) the outcome is not so severe. Instead, the introduction of LED panels offers new possibilities for automated lighting calibration, now including the LED panels as a light source \cite{nila2022implication, payne2022openvpcal, HELZLE2023575}. For example, \citet{legendre2022jointly} treats the panel lighting as ambient light, producing an example result in Figure~\ref{fig:legendre2022example} 

\cite{legendre2022jointly} accomplishes this by investigating a set of pre-correction (kernel) matrices applied to inner and outer LED wall frustums. Then with a separate post-correction matrix applied to the in-camera image, pixels are re-maped to the desired/expected color schema. The pre-correction matrices are solved by comparing and minimizing the difference between the expected LED emission spectral sensitivity functions and the resulting sensitivity function captured by the camera. This optimization is accomplished for a given color chart\footnote{A color chart is used to map shades and tints of red, green and blue, where each square in a chart is a different shade.}. Testing on a variety of color charts showed near-optimal results, though resulting errors are limited by the size of the pre- and post-correction kernels. \citet{legendre2022jointly} highlights issues resulting in de-saturation of skin color\footnote{The authors propose further testing on a larger spectrum of skin colors.} and fabrics; which is reasoned by the restrictive color schema from LED panel lighting. We provide a deeper technical explanation of this method in Appendix \ref{app: ICVFX-legendre}.

Similarly, \citet{smith2021smpte} looks at correcting dynamic in-camera-frustum hue changes, though using simpler color transforms. Additionally, \citet{debevec2022hdr} investigates a method for HDR-Image (HDRI) lighting reproduction, going from virtual HDRI setting to a LED volume setting. This essentially inverts the classical IBL problem and is approached by dilating pixels above a given threshold to meet constrains on local average pixel values displayed displayed on a virtual LED wall.

Overall, the use of LED paneling in production is still relatively new. Aside from discussing lighting and colorization, the IC landscape outside of this paradigm is undefined \cite{kavakli2022virtual}. Subsequently, we have aggregated a set of non-scientific resources, in Table~\ref{tab:VSP pointer}, which contribute to cinematographic discussions concerning the use of LED panels. We additionally categorized the list to provide further clarity on subject materials.

\begin{table}[ht]
\caption{Online Resources containing articles, documentation and video-based discussions on virtual stage/volume film production.}
    \centering
    \begin{tabular}{p{0.3\linewidth} | p{0.6\linewidth}}
        Topic & Papers \\ \hline
         General Discussions & \citet{nila2022implication, kavakli2022virtual, kadner2021, pires2022survey, kadner2019virtual, hendricks2022filmmakers, australian2021youtube}\\
         Workflows/Pipelines & \citet{chambers2017large, lux2021youtube} \\
         Motion Tracking & \citet{drew2020, pixel2021} \\
         Actor Imersion & \citet{bennett2020immersive, leane2020}
    \end{tabular}
    \label{tab:VSP pointer}
\end{table}

\subsubsection{Camera Calibration and Localization}\label{sec: camera calibration}
\begin{figure}
    \centering
    \includegraphics[width=\linewidth]{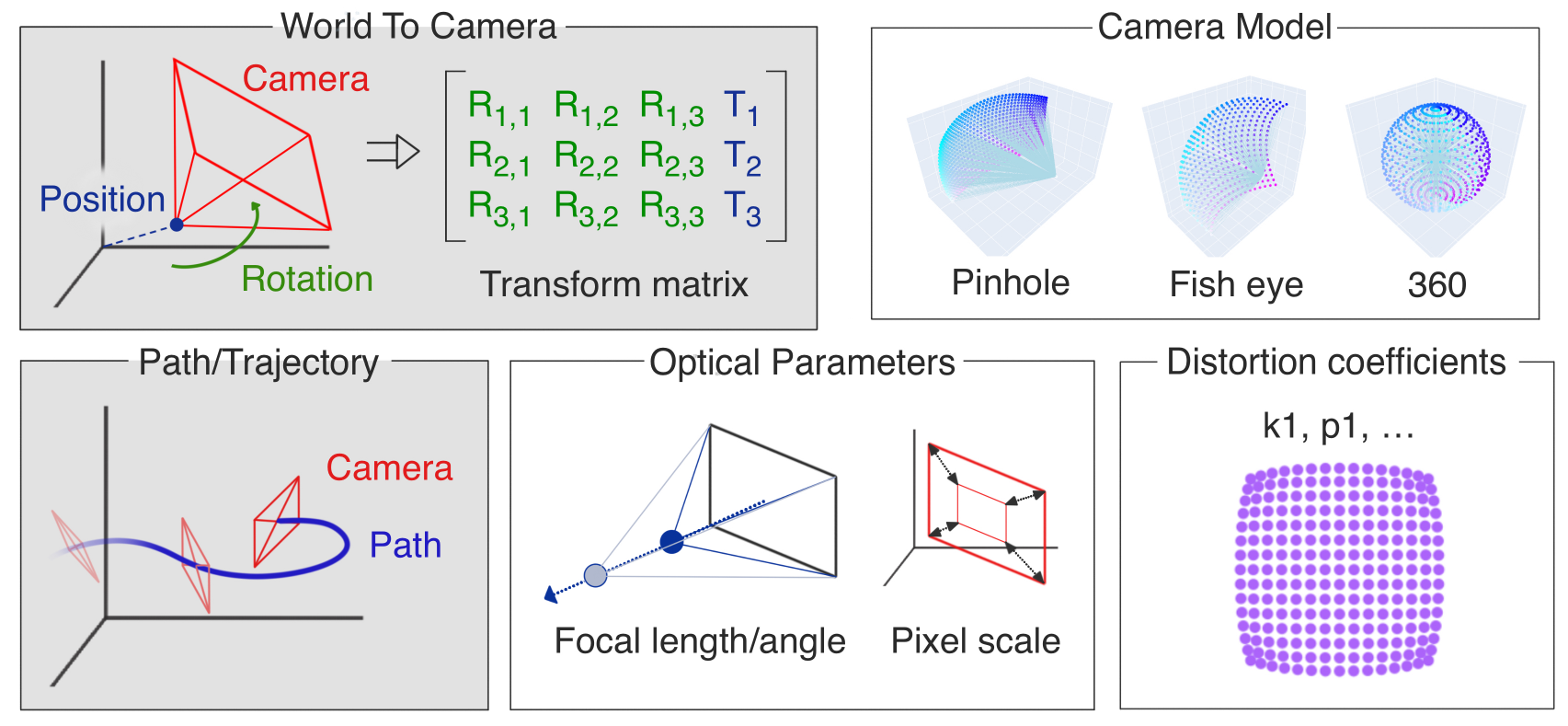}
    \caption{\textbf{Visualizing the parameters involved in camera calibration:} \textit{Extrinsic} parameters are shown in grey. \textit{Intrinsic} parameters are shown in white.}
    \label{fig: camera.parameter}
\end{figure}

The objective of camera calibration and localization (or \textit{pose estimation}) is to map a 3-D world onto a 2-D image plane. This involves modeling camera \textit{intrinsic}/\textit{inertial} parameters and \textit{extrinsic}/\textit{external} parameters \cite{Zhang2014camera} using real image data. An illustration of the relevant parameters to be modeled is shown in Figure \ref{fig: camera.parameter}. The intrinsic parameters define the cameras model, focal length, and lens distortions. The extrinsic parameters model the camera transform matrix for each pose as well as the path for moving shots.

This is a popular problem that concerns the general use case of a single camera \cite{qi2010review, long2019review, remondino2006digital} as well as specific use cases, such as surgical monitoring \cite{qi2010review, obayashi2023multi, koeda2015depth}. Regarding IC, published work relates to several applications including aerial photography \cite{bonatti2019towards, pueyo2022cinempc}, photogrammetry \cite{luhmann2016sensor, clarke1998development, fraser2006zoom}, image-based 3-D reconstruction \cite{chen2022structure, truong2023sparf, xian2023neural, moreau2022lens} and underwater filming \cite{shortis2007review, capra20153d, ma2023combined, massot2015optical, zhou2023underwater}. The two practical uses for this in cinematography are automated camera control and real 3-D reconstruction.

For automated camera control, the research landscape leans towards controlling the extrinsic parameters \cite{salvi2002comparative,remondino2006digital}. However, there exists work bridging extrinsic and intrinsic parametric control to satisfy the cinematographer. For example, \citet{pueyo2022cinempc} present CineMPC which searches for an optimum trajectory (for a drone) and camera angle using a \textit{model predicted control} (MPC) framework. MPC \cite{camachomodel} achieves a process output such as a camera action by considering future time instances/\textit{horizons} and minimizing the cost of selecting different actions. CineMPC models a finite horizon which is continually displaced until all actions cease and constrains the objective function to be flexible to different camera configurations and visual aesthetics. The authors curate mathematical expressions to account for composition, depth of field and canonical shots. Thus, the subsequent cost functions can be considered naive approximations of canonical cinematographic style.

While there exists a lot of work on camera calibration for automated control, the landscape with regard to 3-D reconstruction is more fertile. As discussed by \citet{remondino2006digital}, the general approaches to calibration are \citet{tsai1987versatile}, \citet{heikkila1997four} and \citet{zhang2000flexible} which model distortion intrinsic parameters for a pinhole camera. Nevertheless, there is no universal or flexible automation scheme that achieves this \cite{remondino2006digital}. This is why we still see splintered research on subsets paradigms such as \citet{pitombeira2023influence} which focuses on calibrating the zoom and focus features for scenarios where modeling fixed focus and zoom lenses are infeasible. Or \citet{jasinska2023simple}, which looks at improving geometric stability of videos using SfM of multi-view stereo (MVS) images for photogrammetric reconstruction.

Interestingly, there is work relating to INR and NeRF modeling which addresses the whole problem \cite{zhu2023deep}. The classical model in \citet{mildenhall2021nerf} naively uses COLMAP \cite{schonberger2016structure} whereas recent work has focused on joint-optimization \cite{yen2021inerf, xia2022sinerf, wang2021nerf, truong2023sparf}. While many of these papers improve upon pose estimation for a fixed pinhole camera model, there is work on evolving the lens model as well. For example, \citet{xian2023neural} presents a formulation of a ResNet-based lens distortion model and robust patern-based calibration to provide a thin-lens model equitable for NeRF reconstruction as well as other vision based tasks.

\subsubsection{Neural 3-D Representations of Dynamic Scenes}\label{sec: learning dynamic 3-D representations}
Reconstructing real scenes as 3-D representations presents a host of new solutions to existing  cinematographic problems, such as novel view synthesis for content acquisition, and elevates prior 2-D based paradigms to 3-D. For instance, in Section \ref{sec: general paradigms} we discuss how 3-D reconstruction could be used for evaluating IC research on DA by providing realistic virtual environments, for testing through simulation.

There are two general formulations of this problem:
\begin{enumerate}
    \item \textit{Monocular scenes} captured with a single camera moving around a moving object \cite{pumarola2021d, yan2023nerf, park2023temporal}
    \item \textit{MVS scenes} containing a single action captured using multiple cameras which are often static \cite{sabater2017dataset, broxton2020immersive, li2022neural}
\end{enumerate}

There is a third formulation that has been considered \cite{fridovich2023k}, \textit{forward facing scenes} where a single camera is bounded to a single plane of motion. However this has been weakly adopted as a universal paradigm. For example, K-Planes \cite{fridovich2023k} mimics the general consensus on using normalized device coordinates (NDC) and scene-contraction used in \citet{barron2022mip} to ``hack'' at this problem. However, this offers no ability to render 6 degrees of freedom (6DoF) dynamic video. Doing so would require hallucinating obstructed geometry (the ` ``behind'' of a scene) which introduces a whole new paradigm.

Methods that attend to the general formulations are shown in Table \ref{tab:dynamic nerf comparison}. These methods are not the only concurrent solutions, nonetheless they represent the wide variety of available solutions. 
\begin{table}[ht]
    \centering
    \begin{tabular}{c | c c c c}
        Method & Training Speed & Compact & Explicit & Ancestors \\ \hline
         D-NeRF & D & \textcolor{green}{\checkmark} & \textcolor{red}{\xmark} & NeRF \\
         Dy-NeRF & D & \textcolor{green}{\checkmark} & \textcolor{red}{\xmark} & NeRF \\
         V4D & h & \textcolor{red}{\xmark} & \textcolor{green}{\checkmark} & - \\
         NeRFPlayer & h & \textcolor{red}{\xmark} & \textcolor{red}{\xmark} & Instant-NGP, TensorRF \\
         K-Planes & m/h & \textcolor{green}{\checkmark} & \textcolor{green}{\checkmark} & NeRF-W, Instant-NGP, DVGO \\
         HyperReel & m/h & \textcolor{green}{\checkmark} & \textcolor{red}{\xmark} & TensorRF \\
         NeRF-DS & h & \textcolor{red}{\xmark} & \textcolor{red}{\xmark} & HyperNeRF \\
         HexPlane & m/h & \textcolor{red}{\xmark}  & \textcolor{red}{\xmark} & TensorRF \\
         Tensor4D & h & \textcolor{red}{\xmark} & \textcolor{red}{\xmark} & NeRF-T, D-NeRF \\
         DynIBaR & D & \textcolor{red}{\xmark} & \textcolor{red}{\xmark} & NSFF, IBRNet 
    \end{tabular}
    \caption{\textbf{Overview of Dynamic NeRFs}: \textit{Training Speed} measures the training time where D is days, h is hours and m is minutes. This may be scene-dependent so m/h indicates that training speed varies from minutes to hours. \textit{Compact} indicates whether the models can be trained on a single (commercially available) GPU. \textit{Explicit} indicates where an explicit representation was used to boost inference speed. \textit{Ancestors} indicates the prior work influencing each method}
    \label{tab:dynamic nerf comparison}
\end{table}

There are two traits that currently differentiate research approaches. The first is the proposal of new space-time representations. The second is the manipulation of representations to enhance the learning of temporal elements.

One of the earliest and easiest-to-understand proposals for a new space-time representation is D-NeRF \cite{pumarola2021d}. This method models the deformation (change in position) of volumes w.r.t time. Though intuitive, deformation models present issues when a scene/objects are not continuously in-frame. Instead, methods like K-Planes \citet{fridovich2023k} use low-rank decompositions (e.g., 4-D to 2-D projections) to consistently model scene topology even when objects go unseen for several frames. We discuss D-NeRF and K-Planes in more technical detail in Appendix \ref{app: dynamic nerfs}. To overcome issues with large, slow-to-train dynamic representations,  Gaussian Splatting alternatives such as 4DGS \cite{wu20234d} are preferable. These use the same K-Plane decomposition but apply it to predict both the visual and positional properties of Gaussian blobs, rather than only the visual properties as K-Planes does.

Another interesting solution to the issue of volumetric consistency is key-frame interpolation of static fields. HyperReel exemplifies this by learning the displacement vectors and geometric primitives of a jointly learned key-framed static field. To learn many key-frame fields for a single video, HyperReel builds upon TensorRF due to its compact nature and fast learning ability. Furthermore, as this only provides a discrete set of time-dependent snapshots the authors propose modeling the velocity of volumes at each key-frame. Similarly to the principles of D-NeRF, this enhances the temporal quality of the radiance field representation.

Overall, methods are only moderately capable of modeling dynamic scenes with 6-DoF and usually require additional modification to handle non-generic scenes, such as \textit{forward-facing} \cite{attal2023hyperreel, fridovich2023k}. Despite this, we still find work that tends to production specific needs such as editable NeRFs and GSs' representations \cite{lazova2023control, zheng2023editablenerf, huang2023sc}. For cinematographers this means being patient as we witness increasing interest and contributions in this field.

\subsubsection{Neural 3-D Representations of Humans}\label{sec: human nerf}
modeling non-rigid or deformable geometry, for example a human, is a classical problem for graphics research. This topic is well aligned with research on dynamic NeRFs and broadens human-centered computer vision research to the 3-D case.

Similarly to the prior subsection, the two general problem formulations are multi-view and monocular scenes, which we use to distinguish current research objectives. We can additionally differentiate work by it's reliance on generalizing humans and their poses. HumanNeRF \cite{zhao2022humannerf} exemplifies both of these points as a method that focuses on the multi-view paradigm while generating generalizable human poses. The authors propose using a NeRF to learn the \textit{skinner multi-person linear model} (SMPL) \cite{loper2015smpl}, i.e. the geometry and appearance of an actor. These features are learned and used to train a novel neural appearance blending field. Similarly to D-NeRF, the generalizable NeRF learns the canonical and deformation field of an actor by taking inputs of an SMPL skeleton and pixel aligned features and outputting the color and density of a sample. The appearance blending field refines texture details by accounting for the color of aligned features from neighbouring views. This model performs well, however it requires carefully placed cameras and struggles with new poses.

MonoHuman \cite{yu2023monohuman} overcomes this by using a \textit{shared bidirectional deformation modules} that disentangles forward and backward deformation into rigid skeletal motion and non-rigid motion. Forward deformation regards the transformation from canonical space to a unique observation space, while backward deformation accomplishes the opposite. To guide training for new poses, forward correspondence features at known key frames are selected from an observation bank and visual features are evaluated relative to the features in the new observation space. This improves volumetric consistency between different observation spaces, meaning sequences of actions can be recovered with more confidence. Additionally, the method is less vulnerable to issues with using monocular video as the observation bank can be used to improve reconstruction of new non-rigid actions.

In cinematographic practice, neural human modeling methods relive the accessibility constraints of motion capture (MoCap) suits \cite{wang2022nerfcap, wang2022nemo} - which are currently viewed as the gold standard for MoCap systems. The authors of the neural motion (NeMo) model \cite{wang2022nemo} exemplify this by testing their proposed framework on athletic actions, taken from the Penn Action Dataset \cite{zhang2013actemes}. Like MonoHuman and HumanNeRF, NeMo generalizes motions using multi-view video. Though notably this is achieved by inputting videos of the same action with varying scene conditions, such as different actors and lighting. To handle un-synchronized actions a phase network is introduced as time-based warping function to align poses (i.e. joint angles and translation of motion) in an action sequence. Furthermore, scene-specific feature embeddings handle the visual differences between scenes. Consequently, the phase and instance features are used as inputs to the NeMo network where the outputs define the joint angles and translation of motion - used to render an SMPL model. 

Conclusively, this field of research is popular and highly relevant to IC. While it is currently limited by training and inference time \cite{wang2022nemo, yu2023monohuman} there is potential for its use in applications like MoCap, which may be additionally beneficial for actor-based ICVFX.

\subsubsection{Challenges and Future Work}
With ICVFX, research is limited by access to expensive technologies such as cameras and LED volumes. Considering the novelty of the LED volume production format, we are certain to see more research emerge as more stages are built. \citet{hog2021vfx} echos this by presenting commentary from numerous production supervisors and specialists who demonstrate grand expectations for the technology.

Aside from current research on improving the quality of stage lighting, there are other areas offering potential. LED volume productions (or \textit{virtual stage production} (VSP)) have become popular, relying heavily on pre-production planning; hence there is potential for supporting tasks like pre-production shot visualization, or 3-D stage design.  This area of research aligns well with 3-D capture and modeling research. Real-time 3-D view-synthesis (rendering) could also play a role in in-camera background projection for VSP. This will involve resolving high-resolution real-time rendering problems associated with projecting large live backgrounds. 

With the recent introduction of open-source pipelines for NeRF \cite{tancik2023nerfstudio, li2023nerfacc}, modifying and testing different NeRFs has become trivial. This has been showcased by popular online creatives such as Wren Weichman (a member the \textit{Corridor Crew} YouTube channel) who demonstrates the use of NeRF for numerous CGI and VFX tasks like asset generation and background projection in a video titled ``I try the tech that WILL replace CG one day''\footnote{Accessible here: \url{https://www.youtube.com/watch?v=YX5AoaWrowY}}. This could be elevated for use in an ICVFX pipeline by combining NeRF/GS methods that model backgrounds with real-time automated foreground segmentation techniques. The benefit to this over existing ICVFX solutions is that we no longer need to manually model realistic virtual background and can instead model digital twins of real environments, which is significantly less time consuming and costly. As a current example, the BBC have been investigating NeRFs for synthesizing new shots that are challenging to acquire manually\footnote{Accessible here: \url{https://www.bbc.co.uk/programmes/p0gz4rr0}}. They are able to achieve this in practice due to a well conceived, scene-specific method of capture. However, this demonstrates that there remains a lack of understanding regarding the optimal strategy for capturing media to train NeRFs. Additionally, current methods for evaluating NeRFs using 2-D image-based metrics such as PSNR, SSIM and LPIPS are perhaps not completely reflective of the accuracy of the 3-D fields which we expect a NeRF network to learn \cite{tancik2023nerfstudio, gao2022nerf}. This is specifically true for the dynamic use-case. Hence, this field still requires further work before we can \textit{confidently} select models for large scale commercial productions. It is also worth noting that automated 3-D capture technologies impact on current concerns over the ownership of an actors visual essence. In Section \ref{sec: social.responsibility} we elaborate on the social responsibility of researchers relating to this ethical dilemma.

\subsection{Live Production}\label{sec: liveproduction}
IC research for live production is conditional on the type of environment, types of actions expected to be displayed (e.g., routines or phases in a performance or sport), location of audience members and the focal point for the digital audience. Additionally, challenging environments and scenarios may lead to degraded visual quality, thus real time solutions are required for image correction or enhancement. For example, image analysis from a game of football is challenged by noise arising from spectators or advertisement banners \cite{penumala2019automated, spagnolo2013non, wu2022enhancing}, while water-sports will face problems of distortion from partially submerged participants, light artifacts from surface reflection and noise from turbulent water \cite{wu2022enhancing, host2022overview}. In this section we link work concerning more research on object tracking and human pose estimation and specifically highlight cases that could inform live shot selection or real-time image correction methods.

\subsubsection{Human Pose Estimation}\label{sec: human pose estimation}
Human pose estimation (HPE) is a popular paradigm in computer vision research \cite{badiola2021systematic, sarafianos20163d, zheng2020deep}. We see cases of HPE used in cinematic scene analysis \cite{wang2023jaws, wu2022evaluation} and also in live event broadcasting. For example, in sporting events, analyzing a player's pose may allow us to to forecast a series of entertaining events which could provoke particular cinematographic shots. For sport the major factor that divides research is the environment. We make this delineation prior to reviewing a methods to emphasize that as solutions are heterogeneous\footnote{E.g. anatomical poses will vary between sport.} \cite{badiola2021systematic}. For example, there are special cases where models take-on specific challenges, such as aligning human poses across multiple scenes using 3D pose estimation \cite{wang2023jaws, song2021human}, in a highly dynamic environments \cite{henning2022bodyslam} or for team sports \cite{bridgeman2019multi}. 

\begin{figure}[ht]
    \centering
    \begin{subfigure}{\linewidth}
        \includegraphics[width=\linewidth]{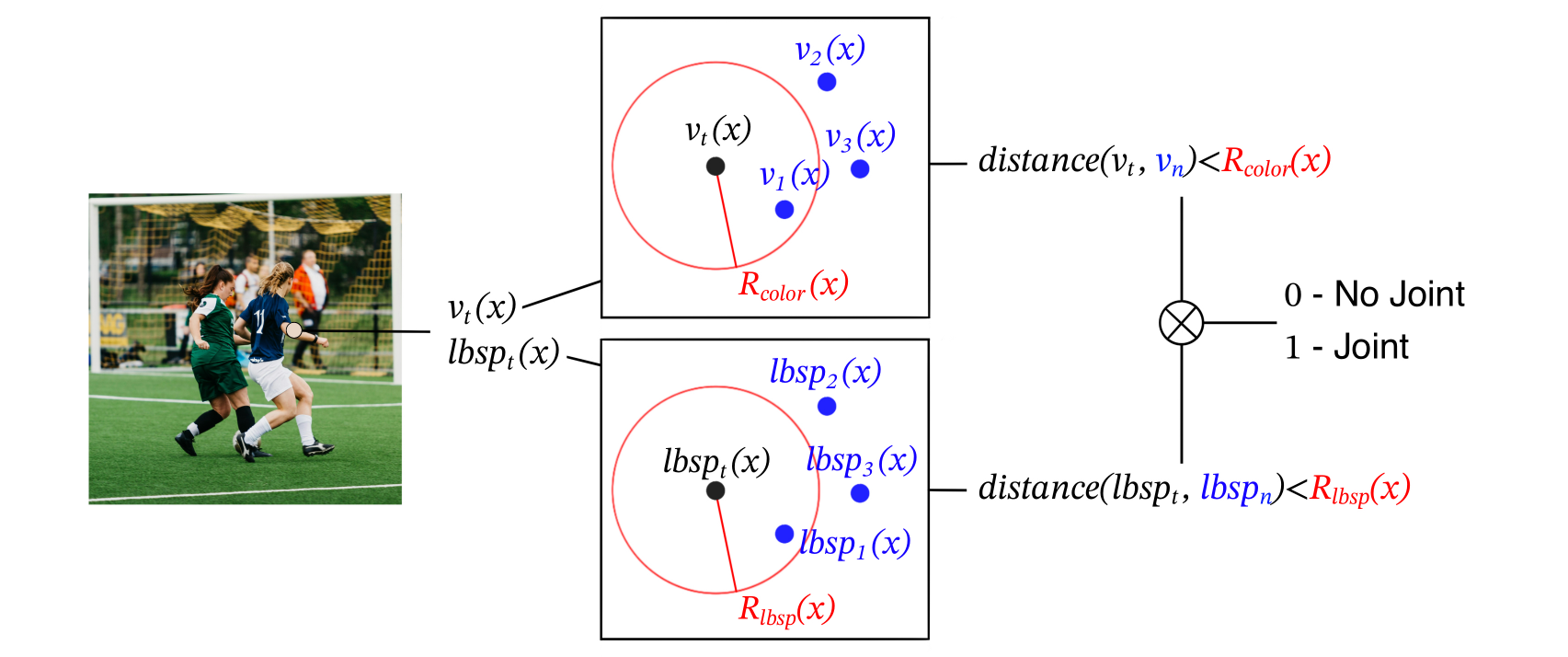}
        \caption{Joint-verification using color features, $v_t$, and the local binary similarity pattern, $lbsp_t$, of pixel $x$ at frame $t$. $R_{color}$ and $R_{lbsp}$ are predefined distance thresholds and $\{v_1, v_2, ..., v_n\}$ and $\{lbsp_1, lbsp_2, ..., lbsp_m\}$ are a set of known color and binary similarity features. The distance function between the pixel's color feature, $lbsp$ feature and features in their respective sets, is evaluated w.r.t the thresholds and the binary result is fused using a logical AND operation to determine where a pixel belongs to a joint.}
    \end{subfigure}
    \begin{subfigure}{\linewidth}
        \centering
        \includegraphics[width=0.7\linewidth]{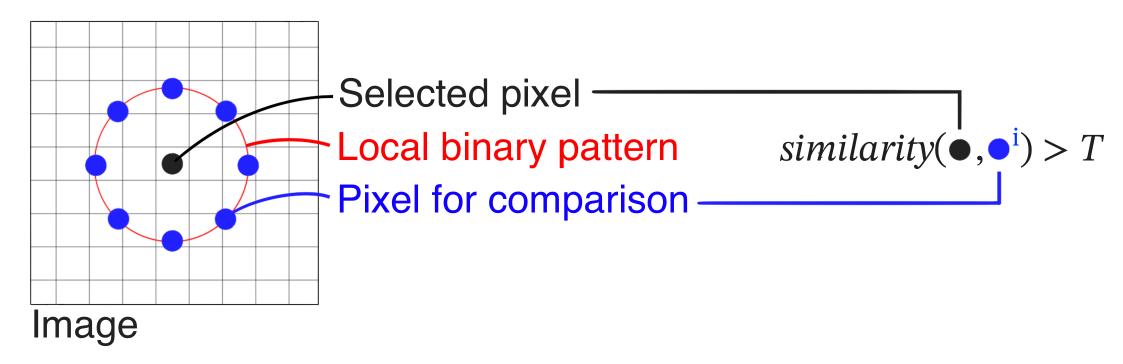}
        \caption{Local binary pattern example, where a binary similarity function evaluates a given pixel with pixels that fall on it's local binary pattern. The similarity is taken w.r.t to a threshold, $T$.}
    \end{subfigure}
    \caption{\textbf{Per-pixel binary joint verification used for identifying pixels for joint classification}}
    \label{fig: hpe.lbsp.examples}
\end{figure}

\citet{hu2021football} looks at real-time football player (single-target) posture estimation for live presentational analysis. To accomplish this, confidence weighted fusion and visual attention is used handle problems with color camouflage and static foreground features, to first identify the target foreground features. Figure\ref{fig: hpe.lbsp.examples} illustrates pixel-based joint verification for identifying key-point target features, using \textit{local binary similarity patterns (lbsp)}. Then a heat map is generated using a ResNet, a stacked hour glass network and deep-layer aggregation (DLA). The DLA collects features from CNN layers to determine the shared features, i.e. it aggregates features at each layer relating to a verified joint and tries to classify what it is (knee, elbow, etc.).For optimization, the model is updated with adaptive strategies relating to the confidence weight of each pixel and their corresponding weighted fusion sum - based on the joint classification. This methods shows significant improvement in joint and posture detection compared to other state-of-the-art methods \cite{wang2013beyond, pishchulin2013poselet, hossain2018exploiting, pavllo20193d}. 

In contrast, \citet{bridgeman2019multi} looks at tracking and analysing multiple targets simultaneously. This method builds upon existing research by \citet{cao2017realtime}. The goal is to construct a 3D (skeletal) pose model using multiple 2D views, while tackling prior issues of long processing time and reliance on appearance-based queues for initiating feature detection. This is achieved by presenting a \textit{greedy} approach to multi-view detection. The approach has three steps: (1) 2D Pose Error Correction is done by flipping incorrectly oriented body parts and dividing by body part, (2) Per-frame 2D pose association determines a consistent label for each body part across multiple views, found greedily by selecting the best pose from a weighted rank, and (3) 3D skeleton tracking uses the 2D labelled poses to generate a 3D skeleton for each individual. The novelty in this paper is it's use of multi-view HPE to verify joint placements in 3D.

With similar ambitions, \citet{song2021human} explicitly improves pose detection methods by using different modules from the VGG11 network for different feature fusion methods. To accomplish this, estimates of features are sampled from the VGG11 network and local features of points are first aggregated. After the image passes through a semantic segmentation network, a segmented target (i.e. a body part) is passed into the feature fusion network. This splices and fuses the RGB features of the segmented image, local point-cloud features and global view features to form a final feature vector which is then reduced to a scalar value for classifying the body part.

\citet{wang2017deep} looks at HPE using fixed aerial cameras. As with previous models, a CNN is used for target (player) detection and a YOLOv2 model is used for pose detection, trained on public aerial image datasets. Processed images undergo further classification of normal and abnormal poses. Conclusively, the model provides an insightful way to detect posture from a common viewing angle for events, though is limited by its ability to detect abnormal poses consequent of a biased dataset - it is challenging to find publicly available datasets pertaining to abnormal poses. It would be interesting to see this developed for real-time application, particularly as wireless video transmission protocols are shifting to improve on wireless-latency and error handling \cite{ahmed2015video}.

\subsubsection{Object Tracking}
As with HPE, object detection and tracking (ODT) has a significant presence in sports broadcasting, for similar reasons. Unlike HPE, solutions are not prescribed on a case-by-case basis. Rather they tackle implicit issues given a set of idiomatic environment characteristics \cite{kaushal2018soft}. For example for football, we could envision a method for detecting a ball in motion within a noisy environment. This translates well to other sports with similar tropes, such as handball.

\cite{kaushal2018soft} presents a comprehensive review of ODT approaches including evolutionary algorithms, conventional neural networks and feed-forward neural networks (FNNs). It is shown that there are suitable approaches for almost any combination of known visual constraints. For example, FNNs are most successful for video streams with irregular illumination change and noise, whilst darker hues and low frame rate are best handled by DL-based CNN classifiers.

Within IC we find two distinct paradigms, (1) singular target ODT and (2) multiple target ODT. (1) is much easier to solve and solutions tend to use YOLO detection. For example, \citet{wu2022enhancing} looks at two sports in particular, swimming and table tennis for \textit{over-the-top} and in-stadium large-screen broadcasting. With the table tennis case study, problems relate to resolving high-speed motion blur for small moving objects to capture the game live in 3-D. While with swimming, problems relate to light-reflection obscuring the cameras vision of swimmers and water occlusion. For table tennis, a YOLOv4 model is used to estimate ball-bounce from a (singular view) video stream and uses this to define a 3-D trajectory. For swimming the solution is a reduced ResNet-50 neural network model acting as the base-network, alongside a modified SiamRPN++ model for swimmer-tracking. To support recognition the camera's view is masked to suppress background noise. \citet{wu2022enhancing} concludes on difficulties with the communication of heterogeneous interfaces between collaborators as well as providing the feasibility of this model in practice. 

As concluded by \citet{buric2018object}, YOLO methods are the fastest to test and fine-tune\footnote{We note only YOLOv2 was tested in this review} and thus are an attractive candidate for practical development. They have also been proven to work well in the presence of  occlusions. The less successful alternative, Mask R-CNN \cite{he2017mask}, uses two networks, one for detecting regions of interest (i.e. a Regional Proposal Network (RPN)) and a deep CNN for determining how likely a group of pixels within a region of interest is associated with an object.

On the other hand, point (2) concerns multiple target ODT which we split into two further sub-objectives: (i) ODT for objects of the same family, and (ii) ODT for objects from different families (such as player and ball detection in \citet{moon2017comparative}). \citet{csah2021review} reviews several popular methods for in-field-sports broadcasting. Conventional approaches detect visual changes using classical image processing methods \cite{deori2014survey, balaji2017survey}. Though, these solutions are naive to several constraints such as (ii). Alternatively, DL-based methods are usually constructed as Fast R-CNNs \cite{girshick2015fast} (ancestor of the Masked R-CNN) and YOLO \cite{pobar2020active,rahmad2019badminton, ren2015faster, hurault2020self}. Yet again, there is also a case to be made for Masked R-CNN \cite{zhang2020mask}. Surprisingly, \citet{csah2021review} finds that conventional methods outrank DL approaches - though not by much. It is reasoned that with a larger number of targets to track, deep CNNs are unable to distinguish lower resolution features whilst conventional methods are totally reliant on high-resolution imagery. In the case of (ii), DL approaches are currently unfeasible as small objects detection presents further issues tied to low resolution feature detection \cite{csah2021review}.

\subsubsection{Challenges and Future Work}
Relative to other forms of production, live broadcasting has a long standing relationship with AI research. For example, the BBC have been involved with supporting AI research in live production settings,  by collecting and freely providing the Old School dataset \cite{Jolly_Phillipson_Evans_2023} to support televised production tasks such as enhanced shot framing, \cite{bbc2022data}. The relationship with industry can be seen through the considerable efforts to elevate the efficacy of pose and object detectors for human targets. However, We have doubts on the current practicality of these models considering their lack of accuracy relative to human ability, although this is well understood by the research community. 
Considering the number of processes involved, there are several other feasible avenues for research. For example, we could extend the supporting infrastructure by investigating different masking strategies for suppressing the general set of noisy surroundings. We could also investigate image detection and correction methods of visual artifacts, such as the reflection of light from a swimming pool \cite{wu2022enhancing}.

Furthermore, a significant quantity of footage is readily accessible online. However researchers must be cognizant of biases in the popularity, gender and racial diversity of sporting events. This indicates a need to source more diverse data-sets for general use.

IC could play a larger part in driving content capture. For example, more research in optimizing shot composition for anticipating events would be of benefit. There may also be interest in stylized shot compositions that present in-game events, for instance, using cinematic style transfer \cite{wang2023jaws} to emphasize praiseworthy shots. Perhaps one day we may be able to watch a sports game in a Tarantio-esque style.

\subsection{Aerial Production}\label{sec: aerial production}
Similar to many other applications of IC, the problem landscape for UAV-based cinematography is generally poorly defined \cite{mademlis2018challenges}. \citet{mademlis2018challenges} produces an interesting review of concurrent challenges, outlining that due to badly defined problem space, lack of technological accessibility as well as roughly drawn legal, ethical and safety constraints, it can be challenging to find a solution which fits the general needs of cinematographers. 

In this section we describe both single drone and multi-drone systems. This is important to consider for cinematographers with tight budgets but also as their technology, capability and complexity  differs significantly.

\subsubsection{Automated Single UAV control}\label{sec: automated single uav}
Aside from budget constraints, the choice to use a single UAV may come down to the size and accessibility of the camera's viewing space. For example a small viewing region around an actor to capture close to mid range shots may only require a single versatile drone. With limited airspace around a subject more drones demands additional constraints to avoid crashes further diminishing the possibility of real-time solutions. Thus there is an area of research on single UAV control referred to as \textit{autonomous ``follow me" quadrotors} \cite{joubert2016towards}. There are a number of vision and sensor-oriented solutions within this domain \cite{naseer2013followme, teuliere2011chasing, lim2015monocular, coaguila2016selecting} as well as drone manufacturers that supply this feature \cite{joubert2016towards}, like the 3DR Solo and DJI Phantom. 

To accomplish this work there are a number of physical and technical limitations to consider. The first is the trade off between online and offline solutions \cite{bonatti2020autonomous, puttige2007comparison}. Online solutions \cite{huang2018act, bonatti2020autonomous} make decisions quickly, responsively and are pragmatic when active elements in a scene move in unpredictable ways, such as an actors movement. Additionally, work exists on physical camera and drone drone modification for improved online and onboard capability \cite{huang2018act}. Offline solutions  can enable solving more complex challenges, such as \textit{swarm robotics} \cite{mademlis2019high}.

The second challenge is the method of detecting actors. This is useful for trajectory planning but also for gimbal control. \citet{bonatti2020autonomous} considers using a single shot detector \cite{liu2016ssd}, such as YOLO9000 \cite{redmon2017yolo9000} and Fast R-CNN \cite{ren2015faster}, for a problem demanding smooth gimbal control and determines that Fast R-CNN is optimal given the speed-accuracy trade off for defining a trajectory. They additionally use MobileNet to perform low-memory feature extraction with fast inference speed. Finally, for indefinite actor tracking they use a KCF \cite{henriques2014high}, taking advantage of the fact that some learning algorithms perform better in the Fourier domain. The KCF tracker relies on \textit{kernel ridge regression} which is a kernalized version of the linear correlation filter, forming the basis for the fastest trackers available \cite{bolme2010visual, bolme2009average}.

Finally the third and most evident challenge is how trajectory planning is handled with regard to certain cinematographic objectives. A tempting option is to adapt cinematographic concepts to mathematical expressions that are subsequently optimized for control \cite{pueyo2022cinempc,joubert2016towards}. Alternatively, \citet{ashtari2020capturing} proposes a method that models the dynamic motion of a human camera operator in real time. The authors follow works on modeling the vertical and lateral displacement of a walking patterns \cite{carpentier2017centre, zijlstra1997displacement} combining approaches into a single routine that additionally considers the rotation of the drone and damping effects to simulate different types of camera equipment. Another approach is to let a reinforcement learning agent control the camera motion. \citet{gschwindt2019can} builds on CHOMP \cite{ratliff2009chomp} to parameterize smooth trajectory planning while using a \textit{deep Q network} (DQN) to lead target shot selection. Two methods are proposed for training the DQN. The first method is a human-crafted reward function; like adapting cinematographic shots to optimization functions, this reward function accounts for the actors presences in a shot, the duration and the shot angle. The second method is human-led observation which rewards the DQN relative to the cinematographers subjective opinion.

Ultimately, the design of autonomous single-camera UAV systems is ambiguous. For cinematographers, this means there are a number of trade offs to consider, such as the complexity and feasibility of using more flexibly systems (e.g., drones with gimbals or "follow me" modes). We agree with the premise set by \citet{mademlis2018challenges}, that work still needs to be done unifying the objectives of autonomous drone systems. However we also believe that a universal paradigm could inhibit varying uses, for example mimicking human walk patterns compared and optimizing shot composition present different objectives and outcomes.

\subsubsection{Automated Multiple UAV control}
For multi-UAV technologies, the limitations of single UAV control are amplified. For example, communication, cooperation, finite bandwidth and safety concerns become more precarious to manage as the number of UAVs increases \cite{mademlis2018autonomous}. This is one reason behind the use of schedulers
\cite{capitan2019autonomous, torres2018multidrone}. Scheduling drones can insinuate two types of tasks, a \textit{swarm} of cooperating drones with a unified purpose and non-cooperative drones \cite{mademlis2019high}. For cinematographers, this is comparative to a drone fleet being used to capture multiple views achieving the same cinematographic objective (for example, a thematically driven system \cite{nageli2017real}) or a fleet used to capture a varying selection of shots for sport or event like filming, where a human or autonomous director makes the final decision over which shot to broadcast \cite{capitan2019autonomous}.

Additionally delineating from single UAV capture, it becomes less plausible to use GPS-based localization for trajectory planning \cite{mademlis2018autonomous}. This is because GPS systems introduce noise during drone localization which perturbs drone formations that demand accurate localization. Furthermore, a similar problem is faced with SLAM-only methods for localization. Hence, an inertial measurement unit (IMU)-GPS-SLAM fused system \cite{mademlis2018autonomous} has been employed. This problem is also shared with works that lie outside of the IC scope \cite{yan2022real, han2022tightly, debeunne2020review}. For example, \citet{yan2022real} looks at a fusing a VO\footnote{Discussed in Section \ref{sec: technological background}} system with IMU integrated on an extended Kalman filter; inspired by \citet{bloesch2015robust}. With the additional curation of a dense 3-D map of the target environment, built using ORB-SLAM \cite{wang2016localization}, automatic robotic localization and control in an agricultural context (i.e. inside a green house and outside in a field) is made possible. Interestingly, the authors experiment on cluttered and non-cluttered environments which is a shared characteristic of production environments - namely the variation of clutter between different sets. 

\citet{nageli2017real} directly address this problem and further highlight challenges with dynamic entities (drones, humans, moving set etc.), using model predicted contour control (MPCC)\footnote{This is related to MPC, previously mentioned.}. Confronted with a highly constrained scenario, the authors simplify the problem using manually defined ``virtual rails" for each drone. This acts as a coarse trajectory guide that drones loosely follow to avoid collisions while achieving one-shot aesthetic objectives. To track and update the state of moving targets a Kalman filter is used. Furthermore, the authors employ an actor-driven view-framing method which can be adjusted in real-time via \textit{general user interface} (GUI) by a cinematographer for varied framing. While not explored in the paper, this could lead to extensions that look at varying the transition between framing inputs at different times to produce aesthetically pleasing transitions. Finally, overcoming the collision constraints, the authors model regions to be avoided as ellipses around a subject as a hard constraint (i.e. high penalization), using slack-variables to indicate when a horizon is foreseeable. Hence, when slack variables are high, the problem is deemed infeasible either due to violating the collision constraints or the computational budget being exhausted. We note that slack variables are commonly used in MPC research to model pareto-optimal solutions.

Overall, the limitations raised by researchers are wide-spread. Since each production scenario is unique and solutions must be robust due to safety concerns, there is no work that currently solves the general paradigm. We share our formalization of the general paradigm with \citet{mademlis2018challenges}, whereby the ideal is a system that can act on a set of high-level inputs from an operator, who is not required to be technically knowledgeable. 

\subsubsection{Challenges and Future Work}
As with the live production setting, aerial cinematography has been strongly influence technological developments. For instance,  AI-controlled drone cinematography is used to capture live events by organizations such as  Quest Films\footnote{More accessible here: \url{https://quest-films.com}}  \cite{john2024cinematographer}. Even commercially available drones are now equipped with AI capture and control features, such the follow me mode in Altair, DJI and Hubsan products.

Following \cite{mademlis2018challenges}, we emphasize the need for discovering a holistic set solutions, building on the wider research literature.
For example, solutions to the UAV localization paradigm canonically involve extensions of SLAM and VO research. With recent developments in camera calibration, consequent to the explosion of NeRF research, we are thus likely to see extended use of visual methods for localization and path generation in the future. Considering this avoids the need for GPS-related hardware, it may result in longer flying time and cheaper budgets which are particularly useful for the multi-UAV paradigm.

Moreover, with new developments in dynamic and human NeRFs, discussed in Section \ref{sec: learning dynamic 3-D representations} and \ref{sec: human nerf}, we see significant potential in training and/or testing automated drones on unforeseen real-world circumstances using digital twins. \citet{boyle2019environment} investigates the use of photogrammetry for training camera operators on single-opportunity shots in live event broadcasting. While the breadth and complexity of test scenarios are limited, we believe NeRFs have potential to fill the gaps, such as introducing more complex human-led interactions. This could also be used to accomplish safety tests so that UAVs, for example, may learn to react to actors disobeying drone safety protocol.

On the other hand, the steady developments in the field of of NeRF technologies could render UAV-based solutions redundant for directly acquiring actor-centric content. Rather, it is likely that UAVs become an intermediary step for NeRF-like content acquisition, as it separates the problem into two steps: (1) UAV image acquisition with the objective of optimizing a NeRF with 6DoF, and (2) content acquisition within NeRF with 6DoF flexibility. This simplifies the current paradigm, which accomplishes localization and path generation constrained by subjective visual objectives.

Alternatively, while this may apply to the general UAV use-case, there are situations that still require attention from IC researchers. For example, the follow-me paradigm implies the use of UAVs for direct content acquisition and relies more on the ability of a drone to move with a visual anchor, like a human target, under cinematographic objectives, like a desired walk pattern or jitter. Considering this is a simpler problem with arguably easier to define objectives, the intervention of NeRF may be unnecessary for specific applications. 

\subsection{Social, Legal and Ethical Challenges} \label{sec: social.responsibility}
The impact of evolving technologies, in particular AI,  on the creative industries is multifaceted and fast changing. In the context of IC, there are several issues that underpin widespread concerns regarding the negative impact of recent AI breakthroughs, such as:
\begin{enumerate}
    \item Replacing actors with human-like AI models \cite{will2023, ringo2023, dawn2023}
    \item Using generative AI and/or 3-D models to produce pornographic content \cite{kat2023, olson2021double}
    \item Using existing art to train generative AI
\end{enumerate}
In the following subsections we expand on these concerns as well as highlight grounded counter arguments where applicable. There are a large number of perspectives to consider, which can be influenced by things like the degree of personal involvement (e.g., researchers involved in generative AI, producers struggling to meet budgets or artists competing against AI for work, etc.), trust in certain sources of information (e.g., news organizations or popular creatives voicing their opinions) or even the spread of mis-information. Hence, we  have tried to remain impartial on these complex topics and encourage readers to form their own opinion.

\subsubsection{Replacing Actors with AI Avatars}
Replacing real actors can be achieved by either modeling a real human or generating a virtual human \cite{will2023}. For actors this can be problematic as they fear abuse or exploitation of existing content containing detailed depictions of themselves. This may also lead to marginalizing real people for virtual celebrity look-a-likes. These points have been emphasized by SAG-AFTRA, a union that represent 160,000 media professionals and was a primary reason for the SAG-AFTRA and Writers Guilds of America strikes that began in 2023. The proposed regulations for AI provided by \cite{sag2023guidelines} primarily deal with consent and compensation by essentially demanding that digital replicas be treated the same as the real media professionals, i.e. ``Residuals paid for use that would normally generate residuals". Digital alteration is also discussed where the need for reasoning (for making cosmetic modification) and transparency (details of the alterations should be conveyed to the union) is emphasized. Moving forward, we believe researchers should consider the ethical implications of producing work that relates to the generation or modification of digital twins or virtual avatars. When work may potentially impact workers, we encourage demonstrating awareness by being transparent about the social impact of future use cases. Importantly, considering the potential future legal ramifications for researchers, more ethical practice could manifest by monitoring and publicly disclosing the use and development of AI datasets and tools as proposed by European Commission, \cite{eu2024guidelines}.

\subsubsection{AI Generated Pornography}
The adult entertainment industry is afflicted by similar problems, and concerns are amplified because of the widespread accessibility of mature content by minors and the treatment of women in the work place \cite{luke2021, trevor2012}. Notably, 96\% of deepfakes are sexually explicit depictions of non-consenting women \cite{kat2023}. Furthermore there are  profound implications of child pornography \cite{olson2021double}. \citet{kirchengast2020deepfakes} reviews current regulatory mechanisms and legislative powers in the US, concluding that present solutions require thorough critique and serious investigation. This is discussed in consideration to the current level of criminalization, which weakly implicates the parties at fault.

The issues raised above relate to our discussions in Sections \ref{sec: virtual production} and \ref{sec: liveproduction}, specifically where a combination of automated 3-D capture and HPE methods could generate trivial solutions for producing deep-fake pornography.
Proposals to address this include: highlighting awareness, applying due-diligence when collaborating with external partners and getting involved in research that aims at developing countermeasures \cite{ashley2024ars, kyle2024csam}. 
Developing robust countermeasures that are easy to implement would provide significant advantages to researchers who wish to publish state-of-that-art models or datasets without contributing to greater social and ethical problems - such those that stemmed from publishing StableDiffusion \cite{shaunagh2024the}.

\subsubsection{Training Generative AI with Existing Content}

The use of existing art to train deep learning models has the potential to impact every artist associated with the cinematographic pipeline. For example, Drew Gooden, a popular online content creator, discusses the impact on stock footage production in a video essay titled ``AI is ruining the internet'', \cite{drew2024ai}. 
The argument is made that AI stock footage often just replaces existing user generated content, rather than generating new content. As with Adobe Stock, the AI content is typically generated by the platform itself and promoted over the existing user generated content, despite being more costly for consumers and lower quality then the user generated content. Therefore, not only does this practice unfairly discard artists in the production pipeline, it can also lead to higher costs for consumers.

Furthermore, platforms have recently begun changing their terms of service to avoid legal challenges associated with using on-platform content for training in-house AI models. Aside from the ethical issues with personal privacy and art theft, this may have a significant impact on publishing promotional content, like trailers or interviews. Considering the SAG-AFTRA regulations on AI (previously discussed), there may be legal challenges associated with uploading content on these platforms. For academic researchers operating outside of industry, this has little relevance. However, for industrial researchers steps that can be taken to minimize the ethical problems. For example, instead of scraping every videos from Facebook into a single database, a more ethical data collection methodology may focus on offering users the opportunity to opt-in to the data collection (e.g., on a per photo/video basis). This should be a significantly important consideration for researchers who are in positions with decisive power.

\section{Concluding Remarks}\label{sec: conclusion}
We have identified several fields of research that are shaping, or have the potential to shape the cinematography industry in the future:
\begin{enumerate}
\item Experimental productions (Section \ref{sec: general.challenges}) offer a way to test IC technologies in real environments. This could avoid biases that come with testing through virtual simulation. Furthermore, production studios may be more willing to adopt research that has been tested in real environments. Considering the current lack of experimental productions and difficulties that come with external collaboration, mutual  benefits could arise from recruiting student actors or hobbyist film makers for their own experimental productions.

\item Computational language structures (Section \ref{sec:film language}) have the potential to  significantly impact future trends in IC. The rapid advances in LLMs and their use in the public domain\footnote{E.g. generating trailers from text/scripts \cite{midjounreryv6tweet2023, aidistrupttweet2023}} will facilitate  natural language decision making for IC in the future. Although we have yet to see significant research in this area,  areas such as automated directive assistance and automated UAV control are candidates for early adoption.

\item The adoption of new LED volume technologies (Section \ref{sec: icvfx and vsp}) has delivered radical changes in canonical chroma-keying and automated roto-scoping (segmentation) practices, offering a new image-based paradigm for computer vision researchers. The ability to perform in-camera compositing and background VFX provides cinematographers with a tighter control over the end product, although equally presents new challenges such as color misalignment and optimal lighting set-ups.

\item Automated 3-D capture (Section~\ref{sec: learning dynamic 3-D representations} and \ref{sec: human nerf}) has delivered flexible new solutions to problems which were previously solved with sculpting/3-D modeling tools and/or hardware dependent photogrammetric methods. With the current push for compact, fast and accurate representations, it is likely that production-ready NeRFs and Gaussian Splats will become practical soon.

\item Human detection and pose estimation (Section \ref{sec: human pose estimation}) plays a large role in research tied to live broadcasting. Currently, research is heavily focused on detection and scene analysis. There are several cases where this has been applied to aerial cinematography (Section \ref{sec: aerial production}), whereby various visual ``aesthetic'' objectives tied to camera position, angle and trajectory have been proposed. Thus, there may be benefit to linking work from human pose estimation (Section \ref{sec: human pose estimation}) and actor-centric aerial capture (Section \ref{sec: aerial production}) to more general forms of production, such as automated shot composition or automated directive assistance (Section \ref{sec:film language} and \ref{sec:film directive}).
\end{enumerate}

This paper also considers the relationship between various domains. For example, for indoor aerial productions (Section \ref{sec: aerial production}) safety testing is challenging and expensive, though could be resolved by simulating UAV trajectories using real digital twins synthesized with NeRF (Section \ref{sec: human nerf}). There also exists cross-over between automated UAV control and automated shot composition \ref{sec: general paradigms}. For example, we could drive autonomous UAVs using methods for attention optimization to improve immersion, or using methods for shot-style transfer to recapture shots without physical rigs (like mechanical dollies). Both of these relations and others are presented in Figure \ref{fig: taxonomy}, whereby we not only highlight existing relations but potential future links between subfields of IC.
<

Conclusively, the relationship between technical research and film making has been highly productive for creative video production, delivering  cinematic innovations such as the \textit{Toy Story}, \textit{The Mandalorian} and \textit{Avatar} sagas. As IC continues to grow as a significant  influence in film-making practice, we hope that this review will support its use in achieving future creative cinematic goals.

\Urlmuskip=0mu plus 1mu\relax
\bibliography{main}

\newpage
\appendix
\section{Appendix: More \textit{Technical Background}}
\subsection{NeRFs}\label{app: more-nerf-background}

The NeRF model samples volume density, $\sigma$, and color radiance, $\mathbf{c}$, provided a 5-D input comprised of 3-D co-ordinates,  $\mathbf{o}\; \epsilon \; \mathbb{R}^{3}$, plus 2-D viewing direction,  $\mathbf{d}\; \epsilon \; \mathbb{R}^{2}$, which represents the position and viewing direction of a sample in space. A sample can be thought of as either a volumetric line-segment along a ray or a voxel intersected by a ray casted into the scene. Simply put, for each pixel $(x,y)$ in an image a ray vector $\mathbf{r_{x,y}}=\mathbf{o} + t \mathbf{d_{x,y}}$ exists, where $t=0$ represents the focal point of an image, $t=n$ represents the position of the image plane along the ray and $t>n$ represents point samples along a ray where $n$ sets the scalar distance from the focal point to the image plane and can vary w.r.t lens distortion. We refer the reader to the Nerfstudio documentation \cite{tancik2023nerfstudio}\footnote{Accessible online: \url{https://docs.nerf.studio/en/latest/nerfology/model_components/visualize_cameras.html}} which overviews the different types of camera models, sampling schemes and sample representations that are found in the NeRF literature.

The original NeRF paper \cite{mildenhall2021nerf} defines the network as a multi-layered perceptron (MLP) with inputs $(r_i, d_i)$ for each sample in a bounded space $t_{near}<t_j<t_{far}$ and outputs $(c_i, \sigma_i)$. To render samples aggregated along a given ray, \citet{mildenhall2021nerf} proposes Equation~\ref{eq:nerfcolor} where the exponent represents the accumulated transmittance w.r.t to the volumetric density of preceding samples. In practice, Equation \ref{eq:nerfcolor} is numerically approximated using quadrature, in Equation \ref{eq:quadNeRFcolor}, where $\delta_i$ is the thickness of a volume sample along a line segment.
 
\begin{equation}\label{eq:nerfcolor}
    \bm{C}(\mathbf{r}) = \int^{\infty}_{t=0} \sigma(\mathbf{r}) \cdot \mathbf{c}(\mathbf{r}, \bm{d})\cdot \bm{e}^{- \int^{t}_{s=0} \sigma(\mathbf{r}) ds} dt
\end{equation}

\begin{equation}\label{eq:quadNeRFcolor}
    \hat{\bm{C}}(\bm{r}) = \sum^{t_{far}}_{i={t_{near}}} (1 - \exp(-\sigma_i \delta_i)) \bm{c_i} \exp({\sum^{i-1}_{j=t_{near}} - \sigma_j \delta_j})
\end{equation}

Subsequently, a loss function $\mathcal{L}(\bm{C^*}, \hat{\bm{C}})$ is used to optimise the predicted ray color w.r.t to the color of a ground truth pixel, $\bm{C^*}$.

To reduce the influence from spectral bias \cite{rahaman2019spectral} the NeRF maps position and viewing directions to a view-point's \textit{fourier features}, using the encoding $\gamma$ in Equation \ref{eq:nerfourier}, where $k$ is a hyper-parameter defining the dimensionality of the feature vector (the bandwidth of our discretized frequency encoding). It is widely reported in the neural representation literature that coordinate-MLPs struggle to learn high-frequency signal details, hence the need for encoding frequencies using $\gamma$. However there are several studies that discuss alternative MLP activation functions which forgo the need for discretized frequency encoding \cite{sitzmann2020implicit, saragadam2023wire}. For example, \citet{saragadam2023wire} proposes an MLP using wavelet activation (called the wavelet implicit representation representation (WIRE)) while \citet{sitzmann2020implicit} proposes a sinusoidal activation (called sinusoidal implicit representation (SIREN)). These are shown to not only reduce the size of the MLP but also capture a higher-bandwidth of frequencies. However, despite \citet{sitzmann2020implicit} and \citet{saragadam2023wire} showcasing higher quality and faster convergence, they discuss how un-actuated these results are in NeRF research.

\begin{equation}\label{eq:nerfourier}
    \gamma^{k}:\mathbf{p}\rightarrow (sin(2^0 \mathbf{p}), cos(2^0 \mathbf{p}), \cdots, sin(2^k \mathbf{p}), cos(2^k \mathbf{p}))
\end{equation}

\subsection{Dynamic NeRFs}\label{app: dynamic nerfs}
D-NeRF \cite{pumarola2021d} models the deformation field $\Phi(\textbf{x}, t) \rightarrow \Delta\textbf{x}$ where $\Delta\textbf{x}$ is the predicted positional change of a ray-sample relative to a canonical static field $\Phi(\textbf{x}+\Delta\textbf{x}, t) \rightarrow (\textbf{c}, \sigma)$. Relative to global space, this learns an SE(3) transformation. Learning a canonical static space is a robust way of ensuring volumetric consistency with time. Though intuitive, this lends itself to issues when a scene is not continuously in-frame. To counterbalance this, methods such as K-Planes decompose static and dynamic volumes representations. More specifically, K-Planes does this by projecting ray samples (containing $(\textbf{x}, t)$) into 6 feature planes, three representing static space and three representing dynamic space. The inputs are normalized between $[0, N]$, projected onto the feature planes and bi-linearly interpolated among varying scales (i.e. coarse and fine features). To decode the features, attaining $(c, \sigma)$, element-wise multiplication is used to recover a final feature vector which is passed into a feature decoder (for explicit representation) or small MLP (for implicit representation). 

\subsection{Image Correction for ICVFX LED Walls}\label{app: ICVFX-legendre}
\cite{legendre2022jointly} accomplishes image correction for ICVFX using LED walls to project virtual background by first applying two matrices, $M$ and $N=MQ^{-1}$ to the out-of-camera-frustum and in-camera-frustum, respectively, where $M$ and $N$ represents a 3$\times$3 pre-correction matrix. Then applying the post-correction matrix, $Q$, to the final image, where $Q$ represents a 3$\times$3 post-correction matrix that re-maps viewed pixels to the desired/expected color schema. $M$ is solved through matrix calculation, from known LED emission spectral sensitivity functions, i.e. $M = [SL]^{-1}$, where $[SL]$ represents the observed average pixel values from capturing light emitted by the LED panels. $Q$ is found by minimising the squared error between predicted pixel values and target pixel values, using the 3x3 matrix $[SRL]_{j}$. This encodes the spectral modulation and integrates the camera spectral sensitivity functions and LED emission and material reflectance spectra \cite{legendre2022jointly} for a given color chart square, $j$. 
\subsection{Generative AI}\label{sec:mlbackground-autoencoders}
Generative AI concerns various domains, for instance text-to-video synthesis \cite{balaji2019conditional, abohwo2023regis, li2018video} and novel view synthesis for dynamic scenes \cite{li2022neural, pumarola2021d, fridovich2023k}. As the novel view synthesis problem-landscape is very broad and specially relevant, we reserve its discussion for Sections \ref{sec:background 3-D modelling} and \ref{sec: virtual production}. Here we focus on generative AI involving 2-D images and video.

\begin{figure}[ht]
    \centering
    \includegraphics[width=\textwidth]{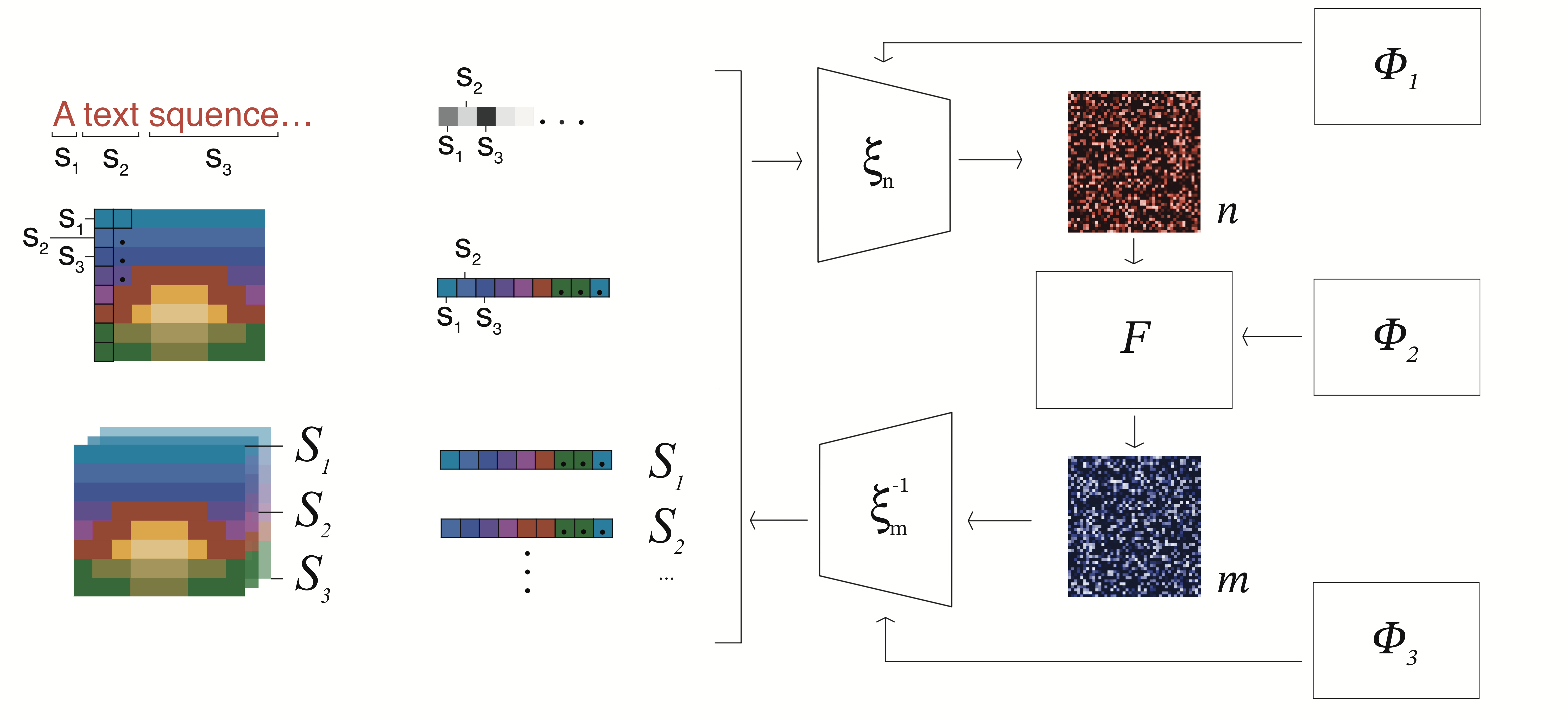}
    \caption{\textbf{\textit{Sequence2sequence} translation.} Text, image and video frame input sequences are encoded into a feature/latent representation and translated/transformed into alternative representations that are decoded to output other tangible representations such as text, images and video frames. $s_i \in S_j$ illustrates how a sequence may be constructed from a set of input data. From top to bottom we give examples for text, image and video sequences. $\xi_n$ encodes the input sequence into a representation, $n$, and a function, $F$, translates the representation into $m$. $\xi_m^-1$ decodes $m$ back into a sequence. $\phi_k$ is a pre-trained network that can be used during the encoding, decoding and/or transformation stage}
    \label{fig:seq2seq}
\end{figure}

The purpose of generative AI varies significantly, dependent on its application. For example, it can involve learning representations for a specific medium, e.g. text or images, and translating/transforming the representation into alternative representations to produce related content e.g. images generated by prompts and vice-versa. Terms such as \textit{text-to-image} or \textit{image-to-text} generation, fall under the broader class referred to as \textit{sequence-to-sequence} translation \cite{venugopalan2015sequence, xu2018youtube, pasunuru2017multi, fajtl2019summarizing}. Figure \ref{fig:seq2seq} demonstrates how abstractions such as text, images and video frames could be sequenced, and simplifies an encoder-decoder pipeline that uses pre-trained networks to inform the encoding, decoding and/or transformation process. Overall, there are four prominent architectures and frameworks that contribute separately to image and video synthesis: (1) encoder-decoder architectures, (2) Generative Adversarial Networks (GANs) \cite{creswell2018generative, goodfellow2020generative, karras2020training}, (3) transformers \cite{vaswani2017attention} and (4) diffusion models \cite{ho2020denoising, zhang2023adding, dhariwal2021diffusion}; discussed below. 

In a GAN (see \cite{jason2019}) a random vector is fed into a \textit{generator} to produce a synthetic representation. Subsequently a comparison is made between the synthetic and real image, using a \textit{discriminator}. The output of the discriminator indicates the level of similarity between the real and generated content, which is subsequently used in a loss function to train both the generator and discriminator. With knowledge of the vectors that successfully generate realistic content, similar vectors can be constructed to produce similar content. This introduces the idea of a \textit{latent space}. Consider a 4-D input vector, $v= [0.2, 0.1, 0.99, 0. 27]$ with values between 0 and 1, that produces an image of a shoe. The values, also called \textit{latent variables}, are used to infer various features. For instance, the first dimension, $v_0=0.2$ may be responsible for modeling features relating to shoe straps and laces, where a $v_0=0$ returns a laced shoe, $v_0=1$ returns a strapped shoe and $v_0=0.5$ returns a shoe with a strap and a lace. Pushing the analogy further, consider the relationship between different latent variables in our input vector. For example, if $v_1$ represents low-top or high-top shoes, then the combination of $v_0$ and $v_1$ will produce high-tops with laces that are tied at the tongue of the shoe and high-tops with laces that are tied at the throat of the shoe. This works so long as the training data set is sufficiently diverse to learn these various features. An acknowledged limitation for this type of GAN is the correlation between the dimensionality of the input vectors (which increases with the number of features/objects to be modeled) and computation cost (including training time, hardware and energy consumed).

To overcome this problem, \textit{conditional} GANs \cite{wang2018high, odena2017conditional, sola2023unmasking, kang2023scaling} have been introduced that condition the random input vectors on context-specific cues, such as object classes or scene descriptors. This is accomplished by inputting the additional information into the generator and discriminator, thus facilitating the generation  of highly diverse content. For example, to generate images of wild animals, the additional information could be the class of animal (e.g. frog, bear and lion). 

For sequence-to-sequence translation, GANs form a part of the translation pipeline, replacing $\xi_m^{-1}$ in Figure \ref{fig:seq2seq}. GANs adapt well to new sequences, although they are less robust at a local-scale; this is where transformers excel. Transformers \cite{lin2022survey, khan2022transformers} are predominantly used to model the relationships between different sequences, replacing $F$ in Figure \ref{fig:seq2seq}, and place greater attention on local context. The \textit{attention} mechanism essentially equates the significance of prior values and their distance from a specific value in a sequence; this process is called \textit{self-attention}. Thus, the attention and contextual value of each point in a sequence is used to generate a representation that is decoded to provide a tangible output. A familiar example is the generative pre-trained transformer (GPT) model \cite{radford2019language, brown2020language}, popularly used for text-to-text and text-to-image translation. For example, if we have a database containing the input question ``Are you ok?" and various outputs, such as ``I am fine" or ``I am sad" the transformer will learn the likeliest response sequence. For example on the local scale, it may be that the majority of answers begin with ``I am", so this is first selected as the start of the output. If a subsequent majority of answers follow this with ``fine" and end the response here, this will make up the second part of the output. There are also cases where context drastically changes the output. For example, a similar question, ``Are you actually ok?", may illicit a meaningful reply if meaningful replies appear more frequently within the training data set. Therefore, more attention might be placed on the word ``actually", triggering a different output.

Ultimately, transformers provide a novel paradigm for text-oriented generative AI, notably machine translation \cite{zhu2022generative, huang2021gpt2mvs}. This is especially important to cinematographers interested in using text-controlled methods either for automated video/image analysis/editing \cite{huang2021gpt2mvs, brooks2023instructpix2pix} or for image and video generation \cite{maniparambil2023enhancing, zhan2021multimodal}. The Stable Diffusion paradigm is closely related to this and  is currently the dominant method for generating images from detailed prompts.

Stable Diffusion embodies a different approach to image and video content generation. Until recently, generative methods, such as GANs, encoder-decoders and autoencoders (described in Figure \ref{fig:seq2seq}) have struggled to provide solutions that offer both low cost (energy, computation, etc.) and fast inference. This is a strength of stable diffusion and a predominant reason for its current success. Stable Diffusion \cite{andrew2023} works by first generating a random N-dimensional tensor - similar to the random vector generated in a GAN. This tensor, together with an encoded prompt, are input into a noise predictor (usually a U-Net \cite{ronneberger2015u, siddique2021u}) which outputs a new tensor predicting the noise in the input tensor. This noise tensor is subtracted from the input tensor, essentially denoising the input.
The new tensor is re-used as an input (now without a prompt) and a new, less noisy, tensor is determined. This process repeats for a fixed number of steps, whereby the final and least noisy tensor is decoded into an image. As the step from a latent space to a less noisy space is a simple operation for a GPU to perform (i.e. a subtraction of two tensors), the generation process is fast. Additionally, with sufficient iterations, the generated image contains negligible noise and is thus of high quality.

Video-based methods using GANs \cite{chen2019mocycle, liang2017dual, liu2021generative}, transformers \cite{wang2021end, selva2023video} and stable diffusion \cite{chai2023stablevideo, karras2023dreampose, ceylan2023pix2video} present a number of underlying challenges, including issues with temporal consistency and a lack of suitable quality assessment metrics. Furthermore, current methods rarely consider camera-based tasks, such as the optimal camera pose or trajectory for generating a video. Considering the current pace of research, we believe it won't be long before this is possible.

In conclusion, generative AI is a powerful tool, likely to offer numerous solutions to production challenges moving forward. However in parallel, researchers must also address ethical issues; for example, relating to the release of code and/or pre-trained models. While many publicly available models (or online application interfaces) incorporate safety measures into their code (like detecting malicious text-based inputs) there are still cases where this is not sufficient for avoiding pornographic content generation \cite{nicky2024microsoft}. For researchers, steps should be taken to raise and act on concerns prior to releasing publicly available code, dataset or pre-trained models.

\end{document}